\documentclass[lettersize,journal]{IEEEtran}
\usepackage{amsmath,amsfonts}
\usepackage{algorithmic}
\usepackage{algorithm}
\usepackage{array}
\usepackage[caption=false,font=normalsize,labelfont=sf,textfont=sf]{subfig}
\usepackage{textcomp}
\usepackage{stfloats}
\usepackage{url}
\usepackage{verbatim}
\usepackage{graphicx}
\usepackage{cite}
\usepackage{multirow}
\usepackage{epsfig}
\usepackage{amsmath}
\usepackage{amssymb}
\usepackage{enumerate}
\usepackage{color}
\usepackage{multirow}
\usepackage{booktabs}
\usepackage{graphicx}
\usepackage{soul}
\usepackage{url}
\usepackage[utf8]{inputenc}
\usepackage{capt-of}

\newcommand{\eat}[1]{}
\begin{document}

\title{KE-RCNN: Unifying Knowledge based Reasoning into Part-level Attribute Parsing}

\author{Xuanhan Wang,
	Jingkuan Song,~\IEEEmembership{Senior Member,~IEEE,}
	Xiaojia Chen \\
	Lechao Cheng,
	Lianli Gao,~\IEEEmembership{Member,~IEEE,}
	and Heng Tao Shen,~\IEEEmembership{Fellow,~IEEE}
	\thanks{\textit{Corresponding author: Jingkuan Song.}}
	\eat{\IEEEcompsocitemizethanks{\IEEEcompsocthanksitem Xuanhan Wang, Jingkuan Song, Xiaojia Chen, Lianli Gao and Heng Tao Shen are with the Center for Future Media and School of Computer Science and Engineering, University of Electronic Science and Technology of China, Chengdu, 611731, China (e-mail: wxuanhan@hotmail.com; jingkuan.song@gmail.com; josonchan1998@gmail.com; lianli.gao@uestc.edu.cn; shenhengtao@hotmail.com). Lechao Cheng is with Zhejiang Lab, Hangzhou, China (e-mail: chenglc@zhejianglab.com).
		}
	}
}

\markboth{Journal of \LaTeX\ Class Files,~Vol.~14, No.~8, August~2021}%
{Shell \MakeLowercase{\textit{et al.}}: A Sample Article Using IEEEtran.cls for IEEE Journals}


\maketitle

\begin{abstract}
	Part-level attribute parsing is a fundamental but challenging task, which requires the region-level visual understanding to provide explainable details of body parts. Most existing approaches address this problem by adding a regional convolutional neural network (RCNN) with an attribute prediction head to a two-stage detector, in which attributes of body parts are identified from local-wise part boxes. However, local-wise part boxes with limit visual clues (i.e., part appearance only) lead to unsatisfying parsing results, since attributes of body parts are highly dependent on comprehensive relations among them. In this article, we propose a Knowledge Embedded RCNN (KE-RCNN) to identify attributes by leveraging rich knowledges, including implicit knowledge (e.g., the attribute ``above-the-hip'' for a shirt requires visual/geometry relations of shirt-hip) and explicit knowledge (e.g., the part of ``shorts'' cannot have the attribute of ``hoodie'' or ``lining''). Specifically, the KE-RCNN consists of two novel components, i.e., Implicit Knowledge based Encoder (\textbf{IK-En}) and Explicit Knowledge based Decoder (\textbf{EK-De}). The former is designed to enhance part-level representation by encoding \textit{part-part} relational contexts into part boxes, and the latter one is proposed to decode attributes with a guidance of prior knowledge about \textit{part-attribute} relations. In this way, the KE-RCNN is plug-and-play, which can be integrated into any two-stage detectors, e.g., Attribute-RCNN, Cascade-RCNN, HRNet based RCNN and SwinTransformer based RCNN. Extensive experiments conducted on two challenging benchmarks, e.g., Fashionpedia and Kinetics-TPS, demonstrate the effectiveness and generalizability of the KE-RCNN. In particular, it achieves higher improvements over all existing methods, reaching around 3\% of $AP^{all}_{IoU+F_{1}}$ on Fashionpedia and around 4\% of $Acc_{p}$ on Kinetics-TPS. Code and models are publicly available at: \url{https://github.com/sota-joson/KE-RCNN}.
	
\end{abstract}


\section{Introduction}
\label{sec:intro}
Part-level attribute parsing refers to localize human body parts and identify their attributes within an image, in which multiple persons and countable body parts with their attributes are expected to be resolved in a unified pipeline. It is a fundamental task in computer vision, as it provides fine-grained human understanding an explainable structure of a person. Optimally addressing this task would greatly support a wide range of human-centric applications, such as human fashion analysis \cite{fashion:Inoue_2017_ICCV,fashion:Mall2019GeoStyleDF,dataset:deepfashion2,dataset:fashionpedia} and human behavior analysis \cite{chen2021baseline,parsing:Pastanet,wang2021technical}.

Over the past decade, we have witnessed tremendous success in instance-level recognition tasks due to the advances in regional convolution neural networks (RCNNs). Numerous RCNN based frameworks such as FPN\cite{obj_det:FPN}, Mask-RCNN\cite{ins_seg:mask_rcnn}, Cascade-RCNN\cite{obj_det:cascadedrcnn} and Transformer based RCNN\cite{swintrans} have been developed, which have substantially pushed forward the state-of-the-art methods in object detection \cite{obj_det:FPN,obj_det:trident,obj_det:libra,obj_det:cascadedrcnn,zhang2020sg,DBLP:journals/tsmc/ChenWCLSLCK22,9091936}, instance segmentation \cite{ins_seg:mask_rcnn,insmask_chen2019hybrid} and human parsing \cite{pose_SunXLWang2019,densepose:ktn,parsing:Wang_2020_CVPR,8972408,9275390}. Inspired by this, recent attempts \cite{dataset:fashionpedia,parsing:Pastanet,chen2021baseline,wang2021technical} directly adopt RCNN based frameworks to support part-level attribute parsing, where successful approaches are derived from object detection models by applying a new branch with attribute prediction head on part-level region features. A well-known example is the Attribute-RCNN \cite{dataset:fashionpedia} extended from Mask-RCNN \cite{ins_seg:mask_rcnn}, in which part-level attributes are identified from local-wise part boxes. However, local-wise part boxes with limit visual clues (i.e., part appearance only) will lead to unsatisfied parsing results, since many part-level attributes are not only decided by part-self but also relevant to others. 

\begin{figure}[t]
	\setlength{\abovecaptionskip}{-0.3cm}%
	\setlength{\belowcaptionskip}{-0.5cm}%
	\begin{center}
		\includegraphics[width=1\linewidth]{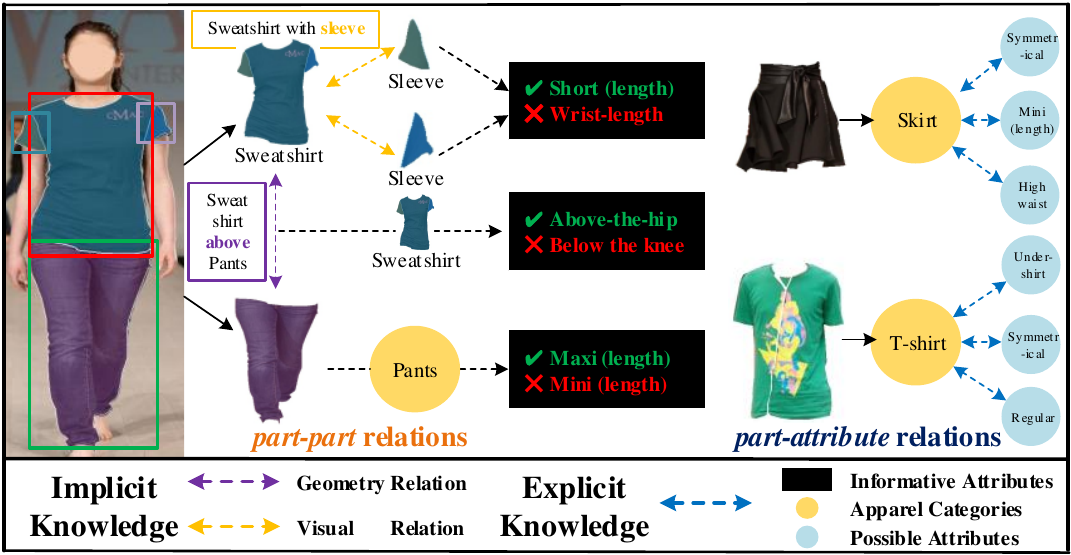}
	\end{center}
	\caption{The implicit knowledge involves rich contexts such as \textit{part-part} visual/geometry relations, and the explicit knowledge refers to commonsense about \textit{part-attribute} relations that are summarized from statistical priors. Jointly considering visual information for a part as well as relational knowledges, i.e., implicit knowledge and explicit knowledge, would greatly benefit part-level attribute parsing. For example, whether a sleeve is short or wrist-length depends on visual relation between sleeve and sweatshirt. With the help of commonsense, many general attributes for a skirt, such as \textit{symmetrical}, \textit{Mini} and \textit{high-waist}, are given without extra process.}
	\label{fig:example}
\end{figure}
To handle above issue, we argue that not only visual information derived from local-wise part boxes but also relational knowledges representing rich clues of a part are needed. There are two reasons behind this: First,
jointly considering part visual information with its implicit knowledge, i.e., parts' visual/geometry context, is crucial for making a correct attribute recognition. For instance, in Fig.~\ref{fig:example} when identifying an attribute of a \textit{``Sweatshirt''} (e.g., \textit{``above-the-hip''}), its visual information is not sufficient enough and we also need its visual contextual information (\textit{e.g.,} $\langle Sweatshirt, upon, Pants\rangle$), which is a geometry relationship to another part \textit{``Pants''}. Other attributes, such as \textit{``Short''} or \textit{``Wrist-length''} of a sleeve, rely on visual relationships between ``sleeve'' with other parts. Secondly, explicit knowledge is essential for understanding parts and their attributes, as it is a piece of wisdom summarized from human practice. Humans are able to identify attributes from complex situations with the help of explicit knowledge. For instance, when identifying attributes for \textit{``Skirt''} in Fig.~\ref{fig:example}, instead of visually observing for an answer, a human tends to intuitively recall an explicit knowledge to infer a set of candidate attributes for \textit{``Skirt''}, such as \textit{``Symmetrical''}, \textit{``Mini''} and \textit{``High-waist''}. As a result, related candidate attributes are provided, and meanwhile numerous irrelevant attributes that are associated with other parts are filtered out. In essential, imitating a human decision process is capable of providing more accurate results but not involving extra computation cost for other unrelated  parts' attributes. 

Motivated by above analysis, in this paper we aim to answer one question: how to utilize implicit/explicit knowledge to enhance part-level attribute parsing. To tackle this, we propose a Knowledge Embedded regional convolution neural network (KE-RCNN), which is a simple yet effective RCNN based framework for part-level attribute parsing. It follows an encoder-decoder design pattern that involves two novel components: (1) Implicit Knowledge based Encoder (IK-En) and (2) Explicit Knowledge based Decoder (EK-De). Specifically, the IK-En is designed to enhance part-level representation by encoding implicit knowledge about $\mathit{part}$-$\mathit{part}$ relational contexts into part boxes, where it smartly decides which $\mathit{part}$-$\mathit{part}$ relations are needed and what contexts to add. After that, the EK-De is proposed to identify attributes from the part-level representation with a guidance of prior knowledge about \textit{part-attribute} relations, which is derived from statistical priors.
With the help of proposed knowledge modeling, the KE-RCNN outperforms state-of-the-art part-level attribute parsing methods. In particular, our KE-RCNN achieves a significant improvement by around 3\% of $AP^{all}_{IoU+F_{1}}$ on Fashionpedia and around 4\% of $Acc_{p}$ on Kinetics-TPS. 

To summarize, our main contributions are three-folds:
\begin{enumerate}[(1)]
	\item We propose an effective part-level attribute parsing method named \textit{Knowledge Embedded RCNN} (KE-RCNN), which identifies part-level attributes by jointly considering visual clues of a part as well as relational knowledge modeling, including implicit knowledge modeling and explicit knowledge modeling.
	\item The KE-RCNN is designed in a play-and-plug fashion and it can be integrated into any two-stage detectors, such as Attribute-RCNN, Cascade-RCNN, HRNet based RCNN and SwinTransformer based RCNN.
	\item Extensive experiments conducted on two challenging benchmarks (i.e., Fashionpedia and Kinetics-TPS) demonstrate the superiority and generalizability of our approach.  
\end{enumerate}

\begin{figure*}[ht]
	\centering
	\setlength{\abovecaptionskip}{-0.3cm}%
	\setlength{\belowcaptionskip}{-0.5cm}%
	\begin{center}
		\includegraphics[width=0.85\linewidth,height=0.4\linewidth]{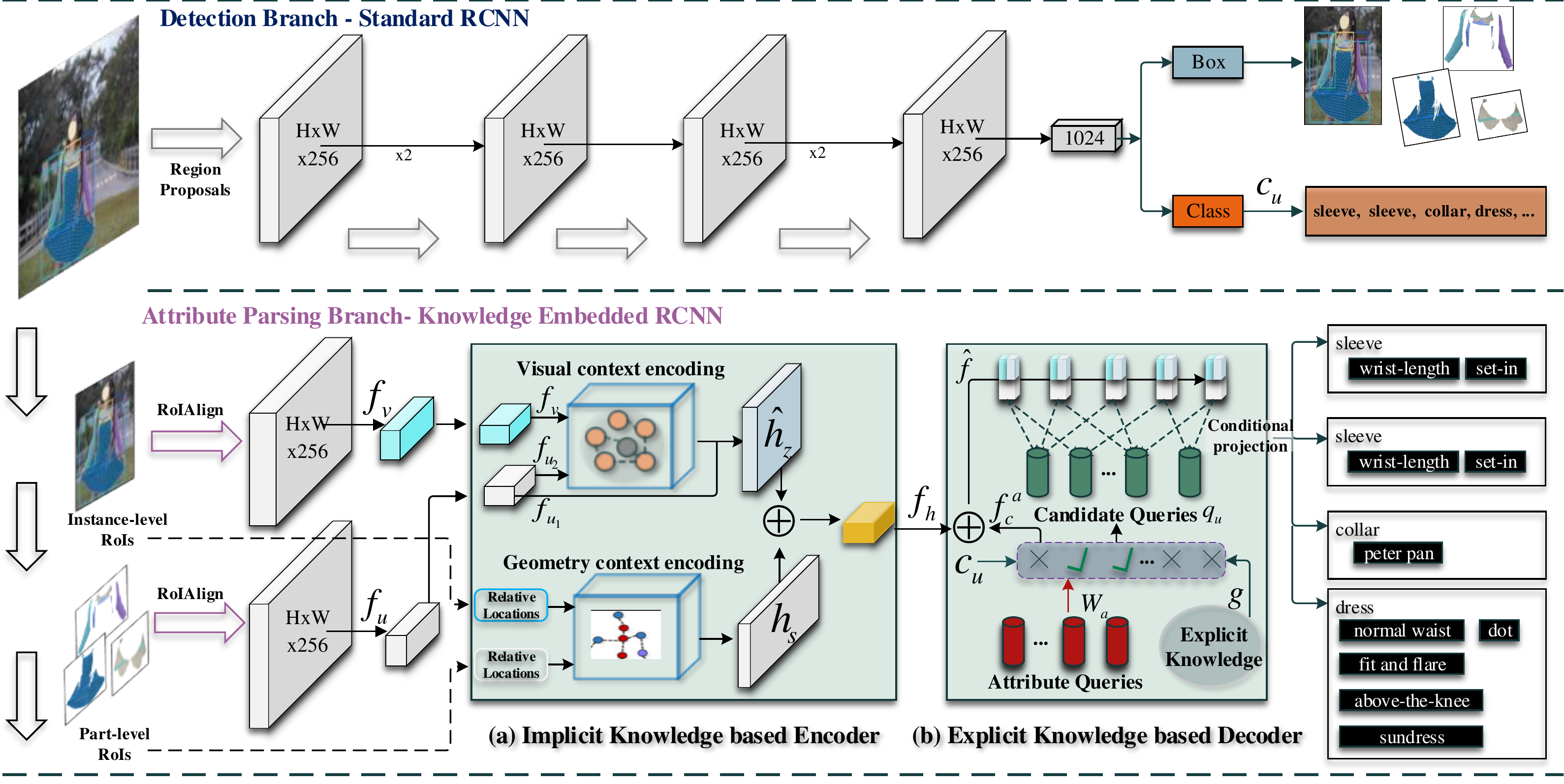}
	\end{center}
	\caption{The overview of the proposed framework: A detection branch with standard RCNN is used to predict bounding boxes for persons and their body parts. For each predicted part, a pair of part-person is built and corresponding region features are extracted by RoIAlign. Then, the KE-RCNN is adopted to identify attributes for each predicted part. In specific, each pair of part-person is firstly fed into implicit knowledge based encoder (IK-En), where visual contexts (top-row of IK-En) and geometry contexts (bottom-row of IK-En) are jointly encoded into part feature. Then, the explicit knowledge based decoder (EK-De) is used to identify attributes of each part by leveraging statistic prior knowledge.}
	\label{fig:framework}
\end{figure*}

\section{Related Work}

\textbf{\textit{Part-level human parsing:}} Traditionally, human parsing aims to segment human bodies into semantic parts. Inspired by the success of RCNN based methods \cite{obj_det:cascadedrcnn,obj_det:trident,ins_seg:mask_rcnn,insmask_chen2019hybrid}, numerous frameworks for part-level human parsing have been developed, which can be categorized into bottom-up, one-stage top-down and two-stage top-down approaches. In general, bottom-up approaches \cite{parsing:Wang_2020_CVPR,parsing:instance-level,pose_AE,pose_cao2018openpose} interpret part-level human parsing as a parsing-then-grouping pipeline, where it firstly predicts instance-agnostic body parts and then groups them into corresponding human instances. Different from them, top-down approaches \cite{densepose:ktn,densepose:parsingrcnn,pose_SunXLWang2019,pose:RSG,pose:stip} firstly detect human instances and then parse each human parts independently, which becomes the mainstream solution in part-level human parsing. Furthermore, the major difference between one-stage and two-stage is whether the human detection branch is combined together with part-level RCNN in a unified manner. 
In a different line of part-level human parsing, recent works \cite{dataset:fashionpedia,parsing:Pastanet,chen2021baseline,wang2021technical} make one more step forward to part-level attribute parsing. In these works, they follow the traditional pipeline and adopt local-wise reasoning by propagating regional visual content only, which may fail since surrounding context of a part is usually required. Instead, our method infers attribute with the help of knowledge modeling, thus improving overall performance of part-level attribute parsing as demonstrated in Section~\ref{exp}.

\textbf{\textit{Knowledge modeling:}} Many works try to enhance deep neural networks by incorporating external knowledges. According to the formation of knowledges, these works can be classified into implicit knowledge based methods and explicit knowledge based methods. Note that implicit knowledge is usually stored in learned models and explicit knowledge is often summarized by human beings. In recent years, knowledge distillation technique and self-attention mechanism are widely used in implicit knowledge based methods \cite{kd:labelencoder,wang2019distilling,kd:mimic,swintrans}. For example, some works \cite{wang2019distilling,kd:mimic} build an implicit knowledge for object detection by training a high-capacity model. Then, they set this trained model as a ``teacher'' and enhance other object detectors with small capacity by distilling implicit knowledge from the ``teacher''. In \cite{swintrans,gberta_2021_ICML}, implicit knowledge about visual context of an object is modeled by an attention mechanism, where only relevant contexts of the object are utilized to facilitate accurate object recognition. 
On the other hand, explicit knowledge based methods adopt statistical priors as the explicit knowledge. In general, the statistical priors are often constructed from large-scale data sources (e.g., Wikipedia or Visual Genome), which record general relations among categories. Therefore, previous works cast those priors as the feature representations and encode them into deep neural networks for addressing the issue of class-imbalance, which facilitates many visual tasks such as scene graph generation \cite{sgg:g2s,sgg:motify,guo2021relation}, object detection \cite{obj_det:reasoning-rcnn} and human parsing \cite{densepose:ktn}. Though impressive, how to explore knowledge to part-level attribute parsing, still remains an open question. As a supplement to them, our method can be viewed as an early attempt to jointly explore implicit and explicit knowledge in the area of part-level attribute parsing. 

\section{Methodology}
In this section, we firstly revisit standard part-level attribute parsing based on Regional Convolution Networks (RCNNs). Then, we give a technical description of our proposed method.

\subsection{RCNN-based Part-level Attribute Parsing}
Given an input image $\mathit{I}$, the goal of part-level attribute parsing is to localize $N$ body parts and identify $C$ attributes for each localized part. Traditionally, a standard pipeline utilizes a backbone network (\textit{e.g.,} ResNet) to project an image to a feature with a size of $H\times W\times D$, where $D$ indicates the number of channes and $\{H, W\}$ denotes spatial size. Then, numerous region proposals are provided by applying a region proposal network \cite{obj_det:rcnn}. Finally, a detection branch, which is a standard RCNN architecture \cite{dataset:fashionpedia} with a box classifier $\varPsi_{B}(\cdot)$, a location regressor $\varPsi_{R}(\cdot)$ as well as an attribute classifier $\varPsi_{A}(\cdot)$, is adopted to simultaneously locate body parts and identify attributes from the region proposals. It is worth noting that attribute parsing depends on the region proposals within the standard pipeline. However, region proposals are low-qualified bounding boxes as they are needed to further refined in RCNN. Therefore, parsing attributes conditioning on region proposals lead to numerous false predictions.

\subsection{Knowledge Embedded RCNN}
In this section, we introduce our proposed method for part-level attribute parsing. An overview of our proposed framework is presented in Fig.~\ref{fig:framework}. Different from standard pipeline that unifies part detection and attribute parsing into one RCNN branch, our method decouples attribute parsing from the standard pipeline and establishes an independent branch, named as KE-RCNN, for attribute parsing. In this setting, we first utilize a standard RCNN to refine region proposals, obtaining final detected boxes of body parts. Then, we apply RoIAlign method \cite{ins_seg:mask_rcnn} to extract part features from final detected boxes instead of low-qualified region proposals. To identify attributes for each detected part, our KE-RCNN first utilizes an Implicit Knowledge based Encoder (IK-En) to enhance the part feature by incorporating \textit{part-part} relational contexts. Then, under a guidance of explicit knowledge about \textit{part-attribute} relations, candidate attribute queries that are relevant to the part are provided. Next, conditioning on candidate attribute queries, the enhanced part feature is further projected to attribute embeddings by applying an Explicit Knowledge based Decoder (EK-De). Finally, a calculated similarity between generated attribute embeddings and attribute queries, is used to identify attributes of the part.

\textbf{Notations.} Before presenting details of our KE-RCNN, we give some notations for clarity. Firstly, we denote $v$ as a detected person and $u$ as one of associative body parts that belong to $v$. Their features extracted by RoIAlign are denoted as 
$f_{v}\in \mathbb{R}^{D \times S_v \times S_v}$ and $f_{u}\in \mathbb{R}^{D\times S_u\times S_u}$, respectively. In addition, $S_v$ and $S_u$ are spatial sizes of features. We denote $c_u\in\mathbb{R}^{N}$ as a predicted categorical distribution about the part $u$, where $N$ is the number of part classes.

\subsubsection{Implicit Knowledge Encoder}

One of our goal is to fully explore implicit knowledge for enhancing part-level attribute parsing. In the following, we discuss the two major components (i.e., visual and geometry context encoding) of our Implicit Knowledge based Encoder (IK-En).

\textbf{Visual context encoding:} Note that each pixel-level person feature $f_v$ represents a set of parts covering the whole person $v$. Thus, a part feature $f_{u}$ can be enhanced by incorporating its visual context relations with other parts by considering relations between $f_{u}$ and $f_{v}$. Following \cite{pami21Res2net}, we start from evenly splitting part representation $f_{u}$ into two subsets, respectively denoted as $f_{u_1} \in\mathbb{R}^{\frac{D}{2} \times S_{u}^2}$ and $f_{u_{2}}\in\mathbb{R}^{\frac{D}{2} \times S_{u}^2}$, which enables our encoder with strong multi-scale feature extraction ability, while maintaining a similar computation cost. Furthermore, each subset has different spatial size and $1/2$ number of channels compared with the person representation $f_v$. In our encoder the $f_{u_{1}}$ is used to represent part visual information, while the $f_{u_{2}}$ is further utilized to encode visual contexts by interacting with $f_v$. Specifically, we first compute an affinity matrix by comparing $f_{u_{2}}$ with $f_{v}$ across all spatial size. Then, each pixel feature in person representation $f_v$ is fused into part representation $f_{u_{2}}$ w.r.t the affinity matrix. Formally, we cast this process as Eq.~\ref{equ.vis_rel_ec}:

\begin{equation}
\begin{array}{lll}
\mathcal{A} & = \sigma((V_1^{T}f_{v})^{T}(Uf_{u_2})) &   \\
h_z & = f_{u_2} + (V_2^{T}f_{v})\mathcal{A} &   \\
\hat{h_z} & = W_z[f_{u_1}, h_z] &   \\
\end{array}
\label{equ.vis_rel_ec} 
\end{equation}
where $V_1\in\mathbb{R}^{D\times\frac{D}{2}}$, $V_2\in\mathbb{R}^{D\times\frac{D}{2}}$ and $U\in \mathbb{R}^{\frac{D}{2}\times\frac{D}{2}}$ are linear matrices that project $f_{v}$ and $f_{u_2}$ into a common embedding space. $\sigma(\cdot)$ is the standard $softmax$ function. $\mathcal{A}\in\mathbb{R}^{S_v^2\times S_u^2}$ is an affinity matrix, which decides what visual context in $f_v$ is needed to propagate to the part. $h_z \in \mathbb{R}^{\frac{D}{2} \times S_{u}^2}$ is an updated part feature that involves relevant visual contexts. $W_z \in \mathbb{R}^{D\times D}$ is a learnable matrix, which linearly fuses the part's visual information $f_{u_1}$ and visual contexts $h_z$ to attain a visually enhanced part representation $\hat{h_z} \in \mathbb{R}^{D\times S_u^2}$.

\textbf{Geometry context encoding:} 
In addition to visual contextual relations, recognizing attributes also benefits from geometry contexts of a part. Specifically, a geometry relation between part $u$ and person $v$ is encoded in their relative locations. Therefore, we represent geometry context of the part $u$ through Eq.~\ref{equ.loc_rel_ec}:
\begin{equation}
\begin{array}{lll}
h_s & = {W_s}(\frac{x_u - x_v}{w_u}, \frac{y_u - y_v}{h_u}, log(\frac{w_v}{w_u}), log(\frac{h_v}{h_u}))^{T} &   \\
\end{array}
\label{equ.loc_rel_ec} 
\end{equation}
where $\langle x_u,y_u,w_u,h_u\rangle$ are coordinates and scales extracted from part region and $\langle x_v,y_v,w_v,h_v\rangle$ are counterpart from person region. $W_s \in \mathbb{R}^{D\times 4}$ is a linear matrix that maps the relative geometry context into a high dimensional vector $h_s\in \mathbb{R}^{D \times 1}$. After that, a part representation $f_h \in \mathbb{R}^{D\times(S_u^2+1)}$ with implicit knowledge (i.e., visual relation and geometry relation) is obtained by simply fusing $h_s$ and $\hat{h_z}$, which is formalized in Eq.~\ref{equ.part_fuse}.
\begin{equation}
\begin{array}{lll}
f_h & = h_s \oplus \hat{h_z}&   \\
\end{array}
\label{equ.part_fuse} 
\end{equation}
where $\oplus$ is a concatenate operation.

\subsubsection{Explicit Knowledge based Decoder} 
In this section, we introduce how to identify attributes of the part by our Explicit Knowledge Embedded Decoder (EK-De). In particular, our key idea is to decode attributes relying on human prior knowledge, which differs from former approaches \cite{dataset:fashionpedia,parsing:Pastanet} that directly apply an attribute classifier $\varPsi_{A}(\cdot)$ on part representations. Compared with directly applying attribute classifier, decoding with human prior knowledge helps to make a correct attribute recognition, as it alleviates adverse effect from decoding irrelevant attributes. Therefore, we embody the EK-De as a conditional projection $\mathcal{J}(\mathcal{X}|G,\mathcal{I},Q)\longrightarrow\mathcal{Y}$, which maps input variable $\mathcal{X}$ to output variable $\mathcal{Y}$ conditioning on $G$, $\mathcal{I}$ as well as $Q$. For clarity, we denote $\mathcal{X}$ as a part representation and $\mathcal{Y}$ as decoded attributes. $G$, $\mathcal{I}$ and $Q$ respectively represent explicit knowledge, part identifier and attribute queries. Next, we present details of each element.

\textbf{Explicit knowledge $G$:} Statistical relations between \textit{part-attribute} pairs provide strong priors to infer an attribute, and it is beneficial to identifying attributes of a part. Therefore, we define explicit knowledge as an undirected relation graph $G:G=<\mathcal{K}_P, \mathcal{K}_A, E>$, where $\mathcal{K}_P$ denotes part categorical nodes and $\mathcal{K}_A$ denotes attribute nodes, $E$ is a set of edges which encode all pairwise relationships between parts and attributes. Furthermore, we build this graph by calculating a frequent statistics matrix $g\in\mathbb{R}^{N \times C}$ from the occurrence among all \textit{part-attribute} pairs, where $C$ is the number of attribute categories. Specifically, we use all relationship annotations and count frequent statistics of each \textit{part-attribute} relation. After counting, each element in $g$ is further rescaled into (0, 1) by a row normalization.

\textbf{Part identifier $\mathcal{I}$:} Categorical distribution $c_u$ indicates probability of each part classes, which is predicted from part detection branch. For simplicity, we directly use it as the part identifier.

\textbf{Attribute queries $Q$:} Parameters $W_{a} \in\mathbb{R}^{D\times C}$ that come from attribute classifier $\varPsi_{A}(\cdot)$ contain global semantic information about attribute categories since it needs to adapt to all attribute embeddings trained from all part samples. Therefore, we use $W_a$ as initial attribute queries. Given $g$ and $c_u$, attribute queries are further filtered, reminding $\hat{C}$ candidate attribute queries of the part $u$. Formally, we cast this process as Eq.~\ref{equ.attr_query}: 
\begin{equation}
\begin{array}{lll}
c_u^\ast & =  {c_u}^{T}g & \\
W_a' & =  W_{a}\circ c_u^\ast & \\
q_u & =  \theta(W_a' | c_u^\ast) &   \\
\end{array}
\label{equ.attr_query} 
\end{equation}
where $c_u^\ast\in \mathbb{R}^{1\times C}$ denotes a weighting vector that decides which attribute is the candidate. $\circ$ denotes the Hadamard product (element-wise broadcast multiplication) and $W_a' \in\mathbb{R}^{D\times C}$ is the weighted attribute queries. $\theta(\cdot\arrowvert \cdot)$ denotes a filtering function that outputs $\hat{C}$ candidate attribute queries $q_u\in\mathbb{R}^{D\times \hat{C}}$ conditioning on $c_u^\ast$, where each attribute query is selected as a candidate if its' corresponding score in $c_u^\ast$ is higher than a predefined threshold value (e.g., 0).

\textbf{Part representation $\mathcal{X}$:} With all conditions, we now build part representation. Except for enhanced part feature $f_h$ yielded from IK-En, parameters $W_{b}\in\mathbb{R}^{D\times N}$ that come from box classifier $\varPsi_{B}(\cdot)$, are utilized to represent semantics of parts. Next, we project the categorical part representation to categorical attribute representation w.r.t a part identifier $c_u$, which is formalized by Eq.~\ref{equ.cls_to_attr}:  
\begin{equation}
\begin{array}{lll}
f_c^a & = \phi({W_c}^T (W_{b}(c_u\circ g))) &   \\
\hat{f} & = f_h \oplus f_c^a & \\
\end{array}
\label{equ.cls_to_attr} 
\end{equation}
where $\phi$ is ReLU nonlinear function and ${W_c}\in\mathbb{R}^{D\times D}$ is a linear transformation matrix. $f_c^a\in\mathbb{R}^{D\times C}$ is the categorical attribute representation, which is dynamically generated from specific part with \textit{part-attribute} priors $g$. $\hat{f}\in\mathbb{R}^{D\times(S_u^2+1+C)}$ is the final part representation that embeds implicit knowledge from $f_h$ as well as explicit knowledge from $g$.

\textbf{Conditional projection $\mathcal{J}$:} In this work, we define projection function as a composition of two independent functions: $\mathcal{J}= \mathcal{\psi}(\mathcal{M}(\cdot))$, where $\mathcal{M}(\cdot)$ is a projection function that maps part representation to attribute representation and $\mathcal{\psi}(\cdot)$ is a decoding function that identifies attributes of a part. Following\cite{vit} , we implement $\mathcal{M}(\cdot)$ based on standard Transformer, as formalized in Eq.~\ref{equ.attr_gen}:
\begin{equation}
\begin{array}{lll}
f & =  MLP(LN(MSA(q_u, \hat{f}, \hat{f}))) &   \\
\end{array}
\label{equ.attr_gen} 
\end{equation}
where $f\in\mathbb{R}^{D\times \hat{C}}$ is the decoded attribute embeddings. $MSA(\mathit{query}, \mathit{key}, \mathit{value})$ denotes standard multi-head self-attention function that decodes $\mathit{value}$ depending on $\mathit{query}$ and $\mathit{key}$. $LN$ is the LayerNorm function for normalizing input feature and $MLP$ is the Multilayer Perceptron that applies nonlinear transformation on input features. It is worth noting that in original Transformer both $\mathit{query}$ and $\mathit{key}$ are derived from same inputs. However, in our model they are derived from two different representations (i.e., $q_u$ and $\hat{f}$), where the former one is dynamically produced according to the particular part $u$. Compared with original version, our model is more flexible and reliable to produce accurate attributes since numerous irrelevant attribute embeddings are removed by Eq.~\ref{equ.attr_query}. 

Next, we define $\mathcal{\psi}(\cdot)$ as a similarity measurement based on Euclidean distance. Therefore, attributes of the part are identified through a similarity matrix calculated between $q_u$ and $f$, as formalized in Eq~\ref{equ.attr_pred}.
\begin{equation}
\begin{array}{lll}
\mathcal{O} & =  \mathcal{P}(\sum\limits_{i=1}^{D} q_u^{i} \circ f^i) &   \\
\end{array}
\label{equ.attr_pred} 
\end{equation}
where $\mathcal{P(\cdot)}$ is the Sigmoid nonlinear function. $\mathcal{O}\in\mathbb{R}^{\hat{C}}$ is the attribute categorical distribution, where each element indicates predicted probability of attribute category.

\section{Experiment}
\label{exp}
In this section, we perform extensive experiments on two challenging benchmarks, one part-level fashion parsing dataset (i.e., Fashionpedia) and one part-level action parsing dataset (i.e., Kinetics-TPS). In the following, we first introduce implementation details. Then, we compare the proposed KE-RCNN with the previous state-of-the-arts on the two tasks. Next, we perform extensive ablation studies to explore the importance design of KE-RCNN.  

\subsection{Implementation Details}
The KE-RCNN is implemented based on OpenMMLab\footnote{\url{https://github.com/open-mmlab}} on an Ubuntu server with eight Tesla V100 graphic cards. We adopt the FPN, which is pretrained on ImageNet, as backbone model unless otherwise stated. We separately train models on the two aforementioned datasets with their respective annotations. Following common practice used in previous works \cite{obj_det:FPN,ins_seg:mask_rcnn,swintrans}, we use SGD solver for optimizing convolution based model and Adam solver for Transformer based model. When training model on Fashionpedia dataset, learning rate is 1e-4 and it is decreased by 10 at the 28-th and 30-th epoch. Besides, the training is stopped at 32-th epoch. For Kinetics-TPS, we train for 12 epochs, starting from a learning rate of 0.02 and decreasing it by 10 at the 8-th and 11-th epoch. A batch size of 16 is used. To provide full details of our approach, our code is made publicly available.

\subsection{Part-level Fashion Parsing on Fashionpedia}

\begin{table*}[ht]	
	\centering
	\caption{Comparisons with representative methods on Fashionpedia dataset. Comparisons with representative methods on Fashionpedia dataset. The symbol ``$\ast$'' means that baseline models are re-implemented by us. Experimental results indicate that decoupling attribute parsing from part detection branch significantly benefits final performances. Besides, replacing standard RCNN with proposed KE-RCNN can further improve the attribute parsing performance by a large margin.}
	\renewcommand\arraystretch{1.3}
	\resizebox{0.9\textwidth}{!}{
		\begin{tabular}{|c|cc|ccccc|}
			\hline
			\multicolumn{8}{|c|}{Without Attribute Parsing Branch} \\			
			\hline
			Settings & Backbone  & Attribute Parsing branch & $AP^{all}_{IoU+F_{1}}$  & $AP^{outerwear}_{IoU+F_{1}}$ &  $AP^{parts}_{IoU+F_{1}}$ & $AP^{50}_{IoU+F_{1}}$ & $AP^{75}_{IoU+F1}$ \\	
			\hline		
			Attribute-RCNN \cite{dataset:fashionpedia}  & ResNet50  & -  &  26.6 & -  & - &  - & -  \\
			Attribute-RCNN \cite{dataset:fashionpedia} & ResNet101  & -  & 28.6 & -  & - &  - & -  \\
			Attribute-RCNN \cite{dataset:fashionpedia} & SpineNet-49& -  & 32.4 & -  & - &  - & -  \\
			Attribute-RCNN \cite{dataset:fashionpedia} & SpineNet-96 & -  & 34.0 & - & - &  - & -  \\
			Attribute-RCNN \cite{dataset:fashionpedia} & SpineNet-143 & -  & 35.7 & - & - &  - & - \\
			\hline
			Attribute-RCNN$^{\ast}$ \cite{dataset:fashionpedia}  & ResNet50  & - &  27.3 & 34.2    & 8.0 & 36.9   & 30.4  \\
			Attribute-RCNN$^{\ast}$ \cite{dataset:fashionpedia}  & ResNet101  & - & 27.9 & 35.0   & 8.0  & 38.1   & 31.4  \\
			Cascade-RCNN$^{\ast}$ \cite{obj_det:cascadedrcnn}  & ResNet50  & - & 29.3 &  37.1 &  8.2  & 39.0  &  32.1  \\
			Cascade-RCNN$^{\ast}$ \cite{obj_det:cascadedrcnn}  & ResNet101  & - & 29.2 &  36.7 &  8.6  & 39.2  &  31.8  \\
			HRNet$^{\ast}$ \cite{pose_SunXLWang2019}  & HRNet-W18  & - & 25.7 & 31.9    & 8.0 & 35.1   & 29.1  \\
			HRNet$^{\ast}$ \cite{pose_SunXLWang2019}  & HRNet-W32  & - & 27.6 & 34.2    & 8.6 & 37.8   & 31.1  \\
			SwinTransformer$^{\ast}$ \cite{swintrans} & Swin-T  & - & 36.2 &   43.0   & 18.6   &  42.2   & 38.0 \\
			SwinTransformer$^{\ast}$ \cite{swintrans} & Swin-S  & - & 37.3 &  44.2   & 18.9   &  45.2  & 39.3 \\
			\hline
			\hline
			\multicolumn{8}{|c|}{With Attribute Parsing Branch} \\			
			\hline
			Settings & Backbone  & Attribute Parsing branch & $AP^{all}_{IoU+F_{1}}$  & $AP^{outerwear}_{IoU+F_{1}}$ &  $AP^{parts}_{IoU+F_{1}}$ & $AP^{50}_{IoU+F_{1}}$ & $AP^{75}_{IoU+F1}$ \\	
			\hline
			\multirow{2}{*}{Attribute-RCNN$^{\ast}$ \cite{dataset:fashionpedia}}    & \multirow{2}{*}{ResNet50}  & Standard RCNN  & 35.1 & 40.5    & 20.5 &  39.8  & 36.6  \\
			&  &  \textbf{KE-RCNN (Ours)}  & \textbf{39.1}  & \textbf{44.2}   & \textbf{26.0}  & \textbf{40.4} & \textbf{38.8} \\ 
			\hline
			\multirow{2}{*}{Attribute-RCNN$^{\ast}$ \cite{dataset:fashionpedia}}    & \multirow{2}{*}{ResNet101}  & Standard RCNN  & 35.8 & 41.6    & 20.5 &  40.7  & 37.5  \\
			&  &  \textbf{KE-RCNN (Ours)} & \textbf{39.9}  & \textbf{44.8}   & \textbf{26.6}  & \textbf{42.3} & \textbf{40.4} \\ 
			\hline
			\multirow{2}{*}{Cascade-RCNN$^{\ast}$ \cite{obj_det:cascadedrcnn}}    & \multirow{2}{*}{ResNet50}  & Standard RCNN  & 36.9 & 42.2    & 22.1 &  41.9  & 38.5  \\
			&  &  \textbf{KE-RCNN (Ours)} & \textbf{41.2}  & \textbf{47.0}   & \textbf{26.2}  & \textbf{42.5} & \textbf{40.7} \\ 
			\hline
			\multirow{2}{*}{Cascade-RCNN$^{\ast}$ \cite{obj_det:cascadedrcnn}}    & \multirow{2}{*}{ResNet101}  & Standard RCNN  & 39.0 &  45.7   & 22.2 &  43.1  & 40.1  \\
			&  &  \textbf{KE-RCNN (Ours)} & \textbf{42.7}  & \textbf{48.8}   & \textbf{26.7}  & \textbf{44.4} & \textbf{42.7} \\ 
			\hline
			\multirow{2}{*}{HRNet$^{\ast}$ \cite{pose_SunXLWang2019}}    & \multirow{2}{*}{HRNet-18}  & Standard RCNN  & 32.7 &   37.0  & 20.1 & 36.7   &  33.2 \\
			&  &  \textbf{KE-RCNN (Ours)}  & \textbf{36.4}  & \textbf{40.4}   & \textbf{25.7}  & \textbf{38.1} & \textbf{36.3} \\ 
			\hline
			\multirow{2}{*}{HRNet$^{\ast}$ \cite{pose_SunXLWang2019}}    & \multirow{2}{*}{HRNet-32}  & Standard RCNN  & 35.2 &  39.5   & 22.8  &  39.9  & 36.3  \\
			&  &  \textbf{KE-RCNN (Ours)}  & \textbf{39.0}  & \textbf{43.1}   & \textbf{27.0}  &  \textbf{41.7} & \textbf{39.7}  \\ 
			\hline
			\multirow{2}{*}{SwinTransformer$^{\ast}$ \cite{swintrans}}    & \multirow{2}{*}{Swin-T}  & Standard RCNN  & 41.0 &  47.1  & 25.8  &  42.6   & 40.5  \\
			&  &  \textbf{KE-RCNN (Ours)}  &\textbf{42.1}  &  \textbf{48.2} &  \textbf{27.3} & \textbf{42.8}  & \textbf{41.2}  \\ 
			\hline
			\multirow{2}{*}{SwinTransformer$^{\ast}$ \cite{swintrans}}    & \multirow{2}{*}{Swin-S}  & Standard RCNN  & 43.5 & 49.9  & 27.6  &  46.3   & 43.5  \\
			&  &  \textbf{KE-RCNN (Ours)}  &\textbf{44.3}  &  \textbf{51.1} &  \textbf{28.1} & \textbf{45.6}  & \textbf{43.7}  \\ 
			\hline
		\end{tabular}
	}
	\label{tab:all_com_stoa_fashion}
\end{table*}

\noindent\textbf{Dataset and metrics.} Fashionpedia dataset \cite{dataset:fashionpedia} is used for evaluating part-level fashion parsing models. It contains 48k images in total, which are collected from Flickr and free license photo websites. It is divided into two subsets: 45623 images for training, 1158 images for validation, respectively. The part-level annotations cover 46 apparel categories, e.g., dress, shorts, leg warmer, and involve 294 attributes, e.g., fit, above-the-knee, regular. Following official settings \cite{dataset:fashionpedia}, we adopt $AP_{IoU+F_{1}}$ to evaluate the performance. It is an extended version of standard detection metric defined in COCO \cite{dataset:coco}, which considers both IoU score for detected part and macro F1 score for predicted attributes of detected part. Based on this, we report standard mean average precision over the validation set: 1) $AP^{all}_{IoU+F_{1}}$ (the mean of box AP scores across all IoU thresholds (ranging from 0.5 to 0.95), all macro F1 scores, and all apparel categories); 2) $AP^{outerwear}_{IoU+F_{1}}$ for outerwear categories; 3) $AP^{parts}_{IoU+F_{1}}$ for garment parts categories; 4) $AP^{50}_{IoU+F1}$ (the mean of box AP scores across all IoU thresholds and all apparel categories with a $F_{1}$ threshold equal to 0.5) and 5) $AP^{75}_{IoU+F1}$ (the mean of box AP scores across all IoU thresholds and all apparel categories with a $F_{1}$ threshold equal to 0.75).

\noindent\textbf{Main Results.} We compare our approach with the state-of-the-art attribute parsing approaches on the Fashionpedia validation set. Specifically, we choose four representative RCNN based models as baselines, including Attribute-RCNN \cite{dataset:fashionpedia}, Cascade-RCNN \cite{obj_det:cascadedrcnn}, HRNet based RCNN \cite{pose_SunXLWang2019} and SwinTransformer based RCNN \cite{swintrans}. In particular, we implement two versions for each baseline model. In the first version, we unify part detection and attribute parsing into one RCNN branch, which is the same as traditional methods \cite{fashion:Inoue_2017_ICCV}. In the second version, we build an independent RCNN branch for attribute parsing, where it identifies attributes from refined detected boxes rather than region proposals. Following common practices \cite{ins_seg:mask_rcnn,dataset:fashionpedia}, we adopt a standard RCNN as the attribute parsing branch in baseline model, where it consists of four consecutive convolutions followed by two fully-connected layers. Tab.~\ref{tab:all_com_stoa_fashion} lists several standard evaluation metrics for different method/backbone pairs. For fair comparisons with baseline models, we report our re-implementation results of them, which are comparable to or higher than those were reported in papers. 

From the results, we find that all traditional approaches based on standard parsing pipeline are significantly improved after decoupling attribute parsing from standard pipeline, where overall improvements achieve at least 6\% $AP^{all}_{IoU+F_{1}}$. For example, the Attribute-RCNN with proposed KE-RCNN outperforms first version of baselines by 11.8\%$\sim$12.0\% $AP^{all}_{IoU+F_{1}}$ when applying different backbones (i.e., ResNet-50 and ResNet-101). On two higher baselines of 29.3\% $AP^{all}_{IoU+F_{1}}$ using Cascade-RCNN and 36.2\% $AP^{all}_{IoU+F_{1}}$ using Swin-T framework, the gains by KE-RCNN are also high, achieving +11.9\% $AP^{all}_{IoU+F_{1}}$ and +5.9\% $AP^{all}_{IoU+F_{1}}$, respectively. It indicates that the performance of attribute parsing increases as identifying attributes from refined boxes, rather than from low-qualified region proposals. The comparison results between two versions of baseline models also support this fact. This is reasonable since attribute parsing depends on part detection. When comparing the KE-RCNN with baseline models based on second version, experimental results in Tab.~\ref{tab:all_com_stoa_fashion} also show consistent improvements (e.g., around 3\% $AP^{all}_{IoU+F_{1}}$) by KE-RCNN at various evaluation metrics, indicating the effectiveness and generalizability of the proposed method. Furthermore, it also demonstrates that implicit and explicit knowledge modeling is a promising direction for attribute parsing.

\subsection{Part-level Action Parsing on Kinetics-TPS}

\noindent\textbf{Dataset and metrics.} For part-level action parsing, we use the Kinetics-TPS dataset\footnote{\url{https://deeperaction.github.io/kineticstps/}} for model evaluation. It contains 3,809 videos in total, which are collected from a subset of Kinetics dataset \cite{dataset:kinetics}. We randomly pick 30\% of training set as the validation set, resulting in 2,686 videos for training and 1,123 videos for validation. Following official setting, we report several metrics on the validation set: 1) $ACC_p$ (the mean of video classification accuracy conditioned on frame-level Part State Correctness (PSC)); 2) $ACC_s$ (the mean of part-level action classification accuracy); 3) $AP^{box}_{part}$ (the mean of box AP scores across all body part categories).

\noindent\textbf{Main Results.} Similar to experiments conducted on Fashionpedia, we adopt Attribute-RCNN, Cascade-RCNN, HRNet based RCNN and SwinTransformer based RCNN as the baseline models. Corresponding results are shown in Tab.~\ref{tab:all_com_stoa_kinetics}. In line with findings from Tab.~\ref{tab:all_com_stoa_fashion}, proposed KE-RCNN outperforms baselines by a margin (+2.2$\sim$4.4 $Acc_{p}$). Based on this, one can conclude that our method performs general improvement on part-level action parsing problem.
\begin{table}[ht]	
	\centering
	\caption{Comparison with representative methods on Kinetics-TPS dataset.}
	\renewcommand\arraystretch{1.3}
	
	\resizebox{0.99\linewidth}{!}{
		\begin{tabular}{|c|c|ccc|}
			\hline
			Settings  & Attribute Parsing branch & $Acc_{p}$  & $Acc_{s}$ &  $AP^{box}_{part}$   \\	
			\hline
			\multirow{1}{*}{Attribute-RCNN \cite{dataset:fashionpedia}}     & Standard RCNN  & 49.1  & 64.0   & 84.9 \\
			(ResNet50)&  \textbf{KE-RCNN (Ours)}   & \textbf{53.5}  & \textbf{69.8}   & \textbf{84.8}   \\ 
			\hline
			\multirow{1}{*}{Cascade-RCNN \cite{obj_det:cascadedrcnn}}     & Standard RCNN  &  49.0  & 63.5   & 85.0   \\
			(ResNet50)&  \textbf{KE-RCNN (Ours)}   & \textbf{53.2}  & \textbf{69.2}   & \textbf{84.2}   \\ 
			\hline
			\multirow{1}{*}{HRNet \cite{pose_SunXLWang2019}}      & Standard RCNN  & 52.3  & 66.9   & 86.6  \\
			(HRNet-W32)&  \textbf{KE-RCNN (Ours)}   & \textbf{54.5}  & \textbf{70.4}   & \textbf{86.2}   \\ 
			\hline
			\multirow{1}{*}{SwinTransformer \cite{swintrans}}     & Standard RCNN  &  52.2  & 67.2   & 86.2   \\
			\multirow{1}{*}{(Swin-T)} &  \textbf{KE-RCNN (Ours)}   & \textbf{56.2}  & \textbf{72.2}   & \textbf{86.8}   \\ 
			\hline
			\multirow{1}{*}{SwinTransformer \cite{swintrans}}    & Standard RCNN  & 54.1  & 68.9  & 87.1  \\ 
			\multirow{1}{*}{(Swin-S)}  & \textbf{KE-RCNN (Ours)}   & \textbf{57.0}  & \textbf{72.6}   & \textbf{87.6}   \\ 
			\hline
		\end{tabular}
	}
	\label{tab:all_com_stoa_kinetics}
\end{table}
\subsection{Ablation Study}
To deeply analyze the proposed method and its components, we conduct extensive ablation studies on Fashionpedia dataset. We choose the decoupling version of Attribute-RCNN with ResNet50 backbone as the baseline model. In the following, we first conduct ablation study to investigate effects of each proposed component. Then, we would like to attain a further insight into implicit knowledge modeling as well as explicit knowledge. After that, we provide a deep analysis of learned models from a visualization aspect.

\noindent\textbf{Ablation studies of each component.} The KE-RCNN consists of implicit knowledge based encoder (IK-En) and explicit knowledge based decoder (EK-De). In this section, we would like to investigate the effect of each component for attribute parsing. Based on this, we build two additional variants of KE-RCNN, where each consists of either IK-En or EK-De.

\eat{Based on the baseline model, we build additional four variants of our approaches for ablation studies. Therefore, we compare five different settings: 
	1) baseline model, which unifies part detection and attribute parsing into one RCNN branch.
	2) \textit{Decoupling}, which decouples attribute parsing from standard attribute parsing pipeline and establishes an independent RCNN branch with four convolution layers followed by two fully connected layers.
	3) \textit{IK-En}, which indicates that establishing independent RCNN branch by adopting proposed Implicit Knowledge based Encoder.
	4) \textit{EK-De}, which indicates that establishing independent RCNN branch by adopting proposed Explicit Knowledge based Decoder.
	5) KE-RCNN, which establishes independent RCNN branch by adopting both \textit{IK-En} and \textit{EK-De}.
	which can be summarized into three conclusions: 1) Decoupling attribute parsing from part detection branch significantly benefits final performances (27.3 vs 35.1). 2) The proposed components are effective to improve attribute parsing, since both of them outperforms the improved baseline by a large margin (9.9 and 11.1). 3) Compared with implicit knowledge modeling, explicit knowledge modeling is more effective to identify attributes, since the EK-De outperforms IK-En by 1.2 $AP^{all}_{IoU+F_{1}}$.
}
The experimental results are reported in Tab.~\ref{tab:abla_study_fashion}, where all models are tested on two benchmarks (i.e., Fashionpedia and Kinetics-TPS). From the results, we have following findings: 1) Both KE-RCNN with IK-En and KE-RCNN with EK-De outperforms standard RCNN on both Fashionpedia validation set (35.1\% vs 37.2\%, 35.1\% vs 38.4\%) and Kinetics-TPS validation set (49.1\% vs 52.2\%, 49.1\% vs 51.7\%), suggesting that incorporating implicit or explicit knowledge facilitates attribute parsing. 2) The KE-RCNN with EK-De shows better performance (i.e., 37.2\% vs 38.4\%) than that of KE-RCNN with IK-En for fashion attribute parsing . However, the results are reversed (i.e., 52.2\% vs 51.7\%) when applying them for action attribute parsing. This suggests that identifying dynamic attributes (e.g., action state) benefits more from implicit knowledges than that from explicit knowledges, while identifying static attributes (e.g., fashion tags) prefers explicit knowledges. 3) Jointly applying IK-En and EK-De brings the best results on both two tasks, suggesting that each component is complementary to each other for attribute parsing.

\begin{table}[t]	
	\centering
	\caption{Component ablation studies on two benchmarks (Fashionpedia and Kinetics-TPS). Investigating the effect of proposed modules.}
	\renewcommand\arraystretch{1.5}
	\resizebox{0.99\linewidth}{!}{
		\begin{tabular}{|c|c|c|c|c|c|}
			\hline
			\multicolumn{6}{|c|}{Part-level Fashion Attribute Parsing (Fashionpedia)} \\
			\hline
			Parsing Branch &  IK-En & EK-De & $AP^{all}_{IoU+F_{1}}$ & $AP^{outerwear}_{IoU+F_{1}}$ &  $AP^{parts}_{IoU+F_{1}}$ \\
			\hline
			Standard RCNN  & - & - &   35.1   & 40.5 & 20.5 \\ \hline
			KE-RCNN  & \checkmark &  &   37.2 & 41.4 & 24.9 \\
			KE-RCNN  &  & \checkmark  & 38.4  & 42.9 & 26.0 \\
			KE-RCNN  & \checkmark & \checkmark & \textbf{39.1} & \textbf{44.2} & \textbf{26.0}  \\
			\hline
			\hline
			\multicolumn{6}{|c|}{Part-level Action Attribute Parsing (Kinetics-TPS)} \\
			\hline
			Parsing Branch & IK-En  & EK-De &  $Acc_{p}$ & $Acc_{s}$   & $AP^{box}_{part}$  \\
			\hline
			Standard RCNN & - & - &  49.1    &  64.0  &  84.9      \\ \hline
			KE-RCNN & \checkmark &  &  52.2   &  67.7  & 85.1    \\
			KE-RCNN &  & \checkmark &   51.7  &  67.7 &   84.5    \\
			KE-RCNN & \checkmark & \checkmark & \textbf{53.5}    &  \textbf{69.8} &  \textbf{84.8}    \\
			\hline
	\end{tabular}}
	
	\label{tab:abla_study_fashion}
\end{table}

\begin{table}[t]	
	\centering
	\caption{Ablation study on Fashionpedia. Investigating the effect of IK-En variants.}
	\renewcommand\arraystretch{1.2}
	\resizebox{1.0\linewidth}{!}{
		\begin{tabular}{|c|c|c|c|c|c|}
			\hline
			RCNN variants & RoI Size & \# Params &  $AP^{all}_{IoU+F_{1}}$ & $AP^{outerwear}_{IoU+F_{1}}$ &  $AP^{parts}_{IoU+F_{1}}$  \\
			\hline
			Standard RCNN & $14\times 14$& 61.1M & 35.1 &  40.5   & 20.5    \\
			ASPP RCNN & $28\times 28$ & 256.0M & 35.8 &   41.0   &   22.2  \\
			IK-En  & $14\times 14$ & 58.9M  & \textbf{37.2}  & \textbf{41.4} &   \textbf{24.9}    \\
			\hline
	\end{tabular}}
	
	\label{tab:abla_study_encoder_fashion}
	
\end{table}

\noindent\textbf{Ablation studies of part extraction.} The global context of a part is the key to attribute parsing. Hence, we investigate various options of part context modeling and compare three different RCNN variants, including: 1) ``Standard RCNN'', where the attribute parsing branch consists of four consecutive convolution layers for generating the part representation; 2) ``ASPP RCNN'', which refers to inserting multiple dilated convolutions into ``Standard RCNN'' for enlarging receptive fields of bounding boxes; 3) IK-En, which denotes the KE-RCNN is composed of proposed IK-En only.
In particular, we choose ``ASPP RCNN'' as one of the options due to its effectiveness of contextual modeling on bounding boxes, as demonstrated in \cite{densepose:ContinuousSurfaceEmbeddings}. For fair comparison, all models adopt one fully connected layer to predict attributes for a part.
\eat{; and 4) $4\times$ IK-En, which denotes the number of IK-En in KE-RCNN is four.}

From experimental results summarized in Tab.~\ref{tab:abla_study_encoder_fashion}, we observe that replacing ``standard RCNN'' with ``ASPP RCNN'' brings minor gains but requires large model size. It suggests that contextual modeling by enlarging visual receptive field has a little effect to attribute parsing, and simply enlarging model capacity has reached performance bottleneck as well. Secondly, the proposed IK-En outperforms ``Standard RCNN'' by 2.1 $AP^{all}_{IoU+F_{1}}$ and requires less parameters. Similar improvements (i.e., 37.2\% vs 35.8\%) are also observed when comparing IK-En with ASPP RCNN. It is worth noting that the standard RCNN applies convolution operations on a part bounding box only, which cannot model global contexts of a part. Instead, the ASPP RCNN can incorporate global contexts outside a part bounding box into the part representation via dilated convolutions, but lacks modeling on relevant contexts of a part. Different from them, the IK-En further filters out irrelevant global visual contexts and encodes geometry contexts into the part representation, thus outperforming both Standard RCNN and ASPP RCNN. Consistent improvements demonstrates the importance of implicit knowledge modeling.
\eat{Third, stacking more IK-En modules brings minor gains of performance, which is in line with the findings explored from the comparison between ``Standard RCNN'' and ``ASPP RCNN''. } 

\begin{table}[ht]
	\centering
	\caption{Ablation study on Fashionpedia dataset. Investigating the effect of part representation.}
	\renewcommand\arraystretch{1.3}
	\resizebox{1.\linewidth}{!}{
		\begin{tabular}{|c|c|c|c|}
			\hline
			Part representation & $AP^{all}_{IoU+F_{1}}$  & $AP^{outerwear}_{IoU+F_{1}}$ &  $AP^{parts}_{IoU+F_{1}}$  \\ \hline
			$\{\hat{h_z}, h_s, f_c^a\}$  &  \textbf{39.1} & \textbf{44.2} & \textbf{26.0}   \\ \hline
			$\{\hat{h_z}, h_s \}$  & 38.5 & 43.0 & 26.3  \\ \hline
			$\{\hat{h_z}\}$ & 38.2 & 42.4 & 26.5  \\ \hline
		\end{tabular}
	}
	
	\label{tab:abla_study_part_rep}
\end{table}

\noindent\textbf{Ablation studies of part representation $\mathcal{X}$.} In KE-RCNN, the part representation is derived from three sources, i.e., visual contexts $\hat{h_z}$, geometry contexts $h_s$ and attribute representation conditioning on predicted part $f_c^a$. In this section, we investigate the effect of each representation and corresponding experimental results are summarized in Tab.~\ref{tab:abla_study_part_rep}. Specifically, we choose the the KE-RCNN with a score of 39.1\% $AP^{all}_{IoU+F_{1}}$ as the baseline, and gradually remove the $f_c^a$ and $h_s$. From the results, we observe that the score of $AP^{all}_{IoU+F_{1}}$ is decreased from 39.1\% to 38.5\% when removing the conditional attribute representation $f_c^a$. It further drops to 38.2\% after removing the geometry contexts $h_s$. Hence, one can conclude that all representations benefit attribute parsing and visual contexts of a part lead to major contribution among them. We conjecture the possible reason is that most attributes rely on visual appearance while only small part of attributes (e.g., above) depends on the geometry relations. Besides, conditional attribute representation $f_c^a$ provides possible attributes of a part, which are needed to be further refined in EK-De.

\begin{table}[ht]	
	\centering
	\caption{Investigating the effect of prior knowledge.}
	\renewcommand\arraystretch{1.2}
	\resizebox{1.0\linewidth}{!}{
		\begin{tabular}{|c|c|c|c|} 
			\hline
			Modifications  &  $AP^{all}_{IoU+F_{1}}$ & $AP^{outerwear}_{IoU+F_{1}}$ &  $AP^{parts}_{IoU+F_{1}}$  \\
			\hline
			Averaging  &  37.9 &  42.5    &  25.1   \\
			Statistics (Fashionpedia) &  39.1  & 44.2 & 26.0   \\
			Statistics (Wikipedia) & 39.6  & 44.3 & 27.0     \\
			\hline
	\end{tabular}}
	
	\label{tab:abla_study_knowledge_fashion}
\end{table}

\noindent\textbf{Ablation studies of explicit knowledge $G$.} In this section, we investigate the important design of explicit knowledge. In particular, the key component of explicit knowledge $G$ is the frequent statistics matrix $g$, where it decides whether an attribute is relevant to a part or not. Hence, we compare three different settings based on modifications of $g$: 1) \textit{Averaging}, where the frequent statistics matrix $g$ is defined as an averaging matrix that propagates all attribute queries without filtering. 2) \textit{Statistics (Fashionpedia)}, where the data source used for calculating the frequent statistics matrix $g$ is based on annotations in Fashionpedia training set. 3) \textit{Statistics (Wikipedia)}, where the data source used for calculating the frequent statistics matrix $g$ comes from all phrases in Wikipedia Corpus\footnote{\url{https://dumps.wikimedia.org/}}. Tab.~\ref{tab:abla_study_knowledge_fashion} lists corresponding results. From the results, we can find that the explicit knowledge is critical, as the performance  decreases by 1.2\% (37.9\% vs 39.1\%) if the explicit knowledge is removed (see 1st and 2nd setting). Furthermore, using explicit knowledge stored in Wikipedia improves the performance from 39.1 to 39.6. This suggests that explicit knowledge is particularly effective for accurate attribute parsing.

\begin{figure}[ht]
	\centering
	\setlength{\abovecaptionskip}{-0.2cm}%
	\setlength{\belowcaptionskip}{-0.5cm}%
	
	\begin{center}
		\includegraphics[width=0.9\linewidth]{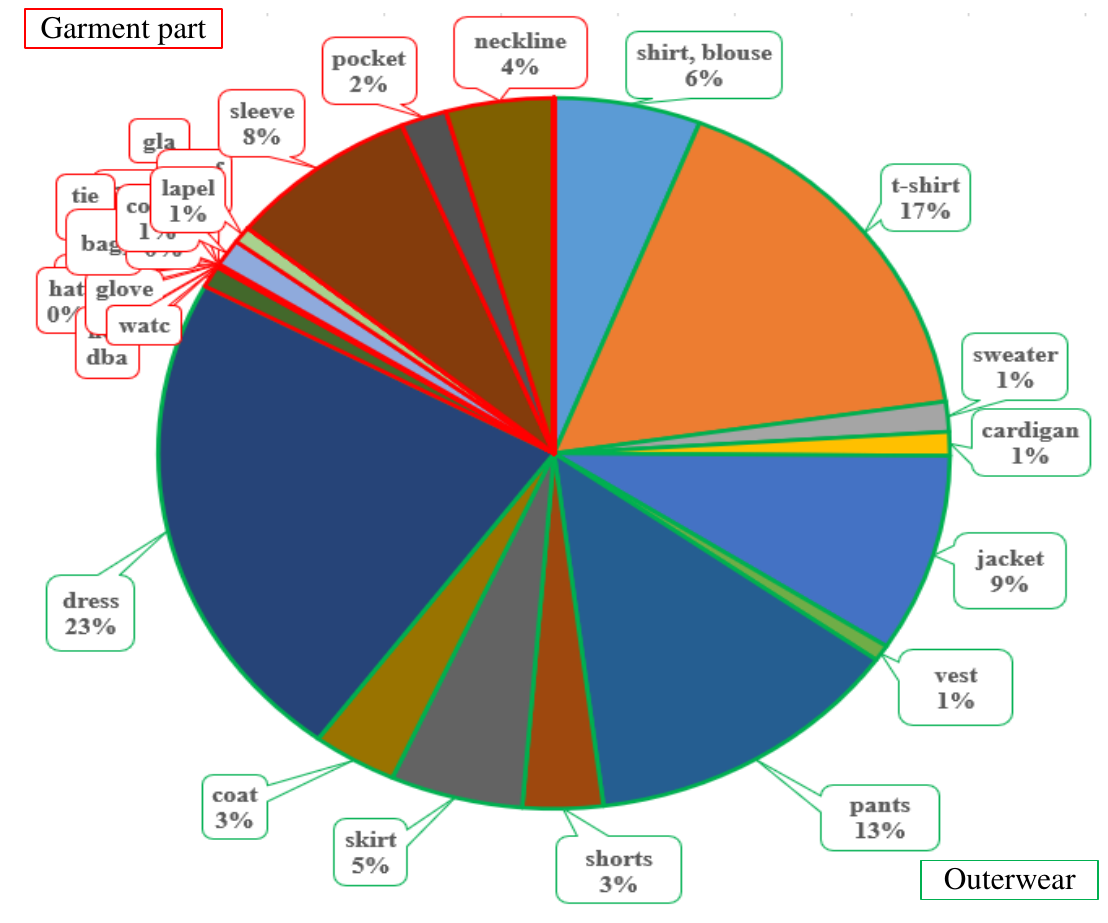}
	\end{center}
	\caption{Annotations distribution of part-level attributes in Fashionpedia presents the characteristic of imbalance: The number of samples for outerwear are much larger than that for garment part.} 
	\label{fig:imbalance-label}
\end{figure}
\begin{figure}[ht]
	\centering
	\setlength{\abovecaptionskip}{-0.2cm}%
	\setlength{\belowcaptionskip}{-0.5cm}%
	
	\begin{center}
		\includegraphics[width=0.99\linewidth]{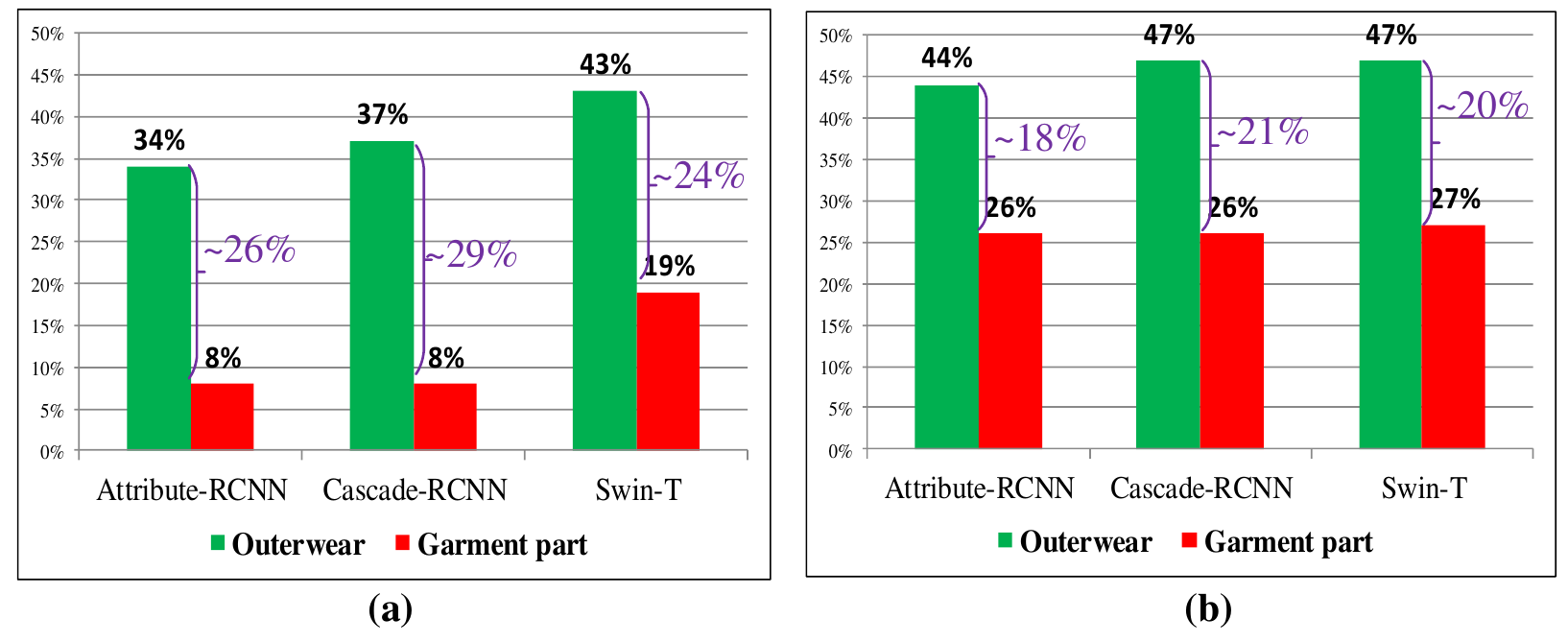}
	\end{center}
	\caption{Performance yielded from baseline models (a) and that from KE-RCNN (b). This suggests that baseline models trained on imbalanced dataset have a bias to major classes (i.e., outerwear), resulting in large performance gap between ``outerwear'' classes and ``garment parts'' classes. With the help of knowledge modeling, the proposed KE-RCNN significantly narrow down the performance gap.}
	\label{fig:imbalance-comp}
\end{figure}
\begin{table}[ht]	
	\centering
	\caption{Investigating the effect of proposed method for attribute parsing of ``hard'' apparel. The symbol ``$\ast$'' means that  model is re-implemented version.}
	\renewcommand\arraystretch{1.2}
	\resizebox{1.0\linewidth}{!}{
		\begin{tabular}{c|ccccc|c}
			\hline
			\multirow{2}{*}{method}          & \multicolumn{5}{c|}{Top-5 Hard Classes} & \multirow{2}{*}{Ave.} \\ \cline{2-6}
			& Collar & Lapel & Sleeve & Pocket & Neckline &  \\ \hline
			Attribute-RCNN$^{\ast}$  &   16.0\%     &  16.9\%     &   43.7\%     &  20.8\%      &   14.2\%       &  22.3\%    \\ 
			w KE-RCNN       &   19.8\%     &  36.8\%     & 46.4\%       &   24.1\%     &   21.0\%       &  29.6\%    \\ \hline
			Cascade-RCNN$^{\ast}$    &     17.5\%	& 17.8\%	 &   46.4\%	 & 23.0\%	& 16.0\%	& 24.5\%       \\ 
			w KE-RCNN       &   19.8\%	& 34.6\%  & 47.3\%	& 25.2\% & 20.5\% & 29.5\%     \\ \hline
		\end{tabular}
	}
	\label{tab:abla_study_fewshot_fashion}
\end{table}
\noindent\textbf{Analysis of learned model.} As demonstrated in \cite{dataset:fashionpedia}, current attribute parsing datasets present the characteristic of long-tail, which seriously hinders model's performance. In particular, parsing models trained on imbalanced datasets are likely dominated by major classes. As shown in Fig.~\ref{fig:imbalance-label}, the number of labeled samples for outerwear in Fashionpedia are much larger than that of garment part, resulting in large performance gap (e.g.,  26\% AP) between them as shown in Fig.~\ref{fig:imbalance-comp}. To investigate whether proposed KE-RCNN suffers from this or not, we evaluate them on ``hard'' cases. In particular, these ``hard'' cases are selected from samples of tail classes. We report evaluation results in Tab.~\ref{tab:abla_study_fewshot_fashion}. From the results, we observe that our method significantly improves two baselines (+7.3 and +5.0) for parsing those ``hard'' classes, indicating that the adverse effect caused by imbalanced issue can be alleviated with the proposed method. It is worth noting that the proposed KE-RCNN identifies attributes of a part with the help of general explicit knowledge, thus improving performance for ``hard'' examples. 

\begin{figure*}[ht]
	\centering
	\includegraphics[width=0.8\linewidth, height=0.4\linewidth]{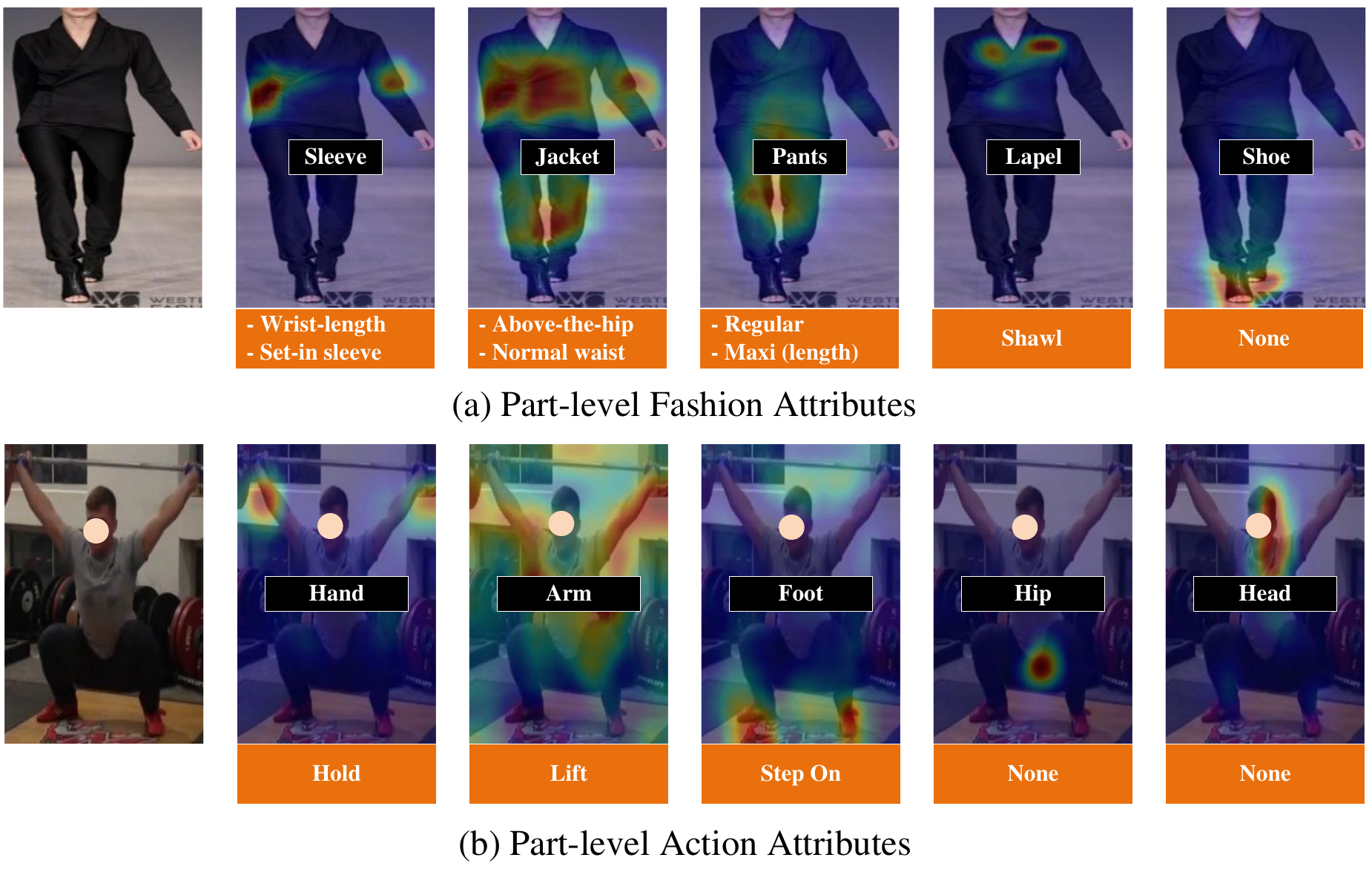} 
	\caption{Investigating what visual contexts the KE-RCNN utilizes to identify attributes.  The first row (a) presents visual contexts for parsing fashion attributes, while the second row (b) displays visual contexts for parsing action attributes. For each example, it visualizes the affinity matrix $\mathcal{A}$, since it decides which part of a person feature is used to identify attributes.}
	\label{fig:vis_att}
\end{figure*}

To further attain insight into the learned model, we visualize the implicit knowledge that the KE-RCNN utilizes to parse attributes of a part. In particular, the affinity matrix $\mathcal{A}$ for each part is visualized, since it decides which part of a person feature is used to identify attributes. The visualization results in Fig.~\ref{fig:vis_att} shows that the KE-RCNN can well focus on relevant contexts for parsing attributes of a part. For example, when identifying ``above-the-hip'' for the \textit{Jacket}, the KE-RCNN focuses on both torso and hip. In terms of action attributes, relevant body parts that perform actions are also captured by the KE-RCNN as well.  

\begin{figure*}[ht]
	\centering
	\includegraphics[width=0.95\linewidth, height=0.4\linewidth]{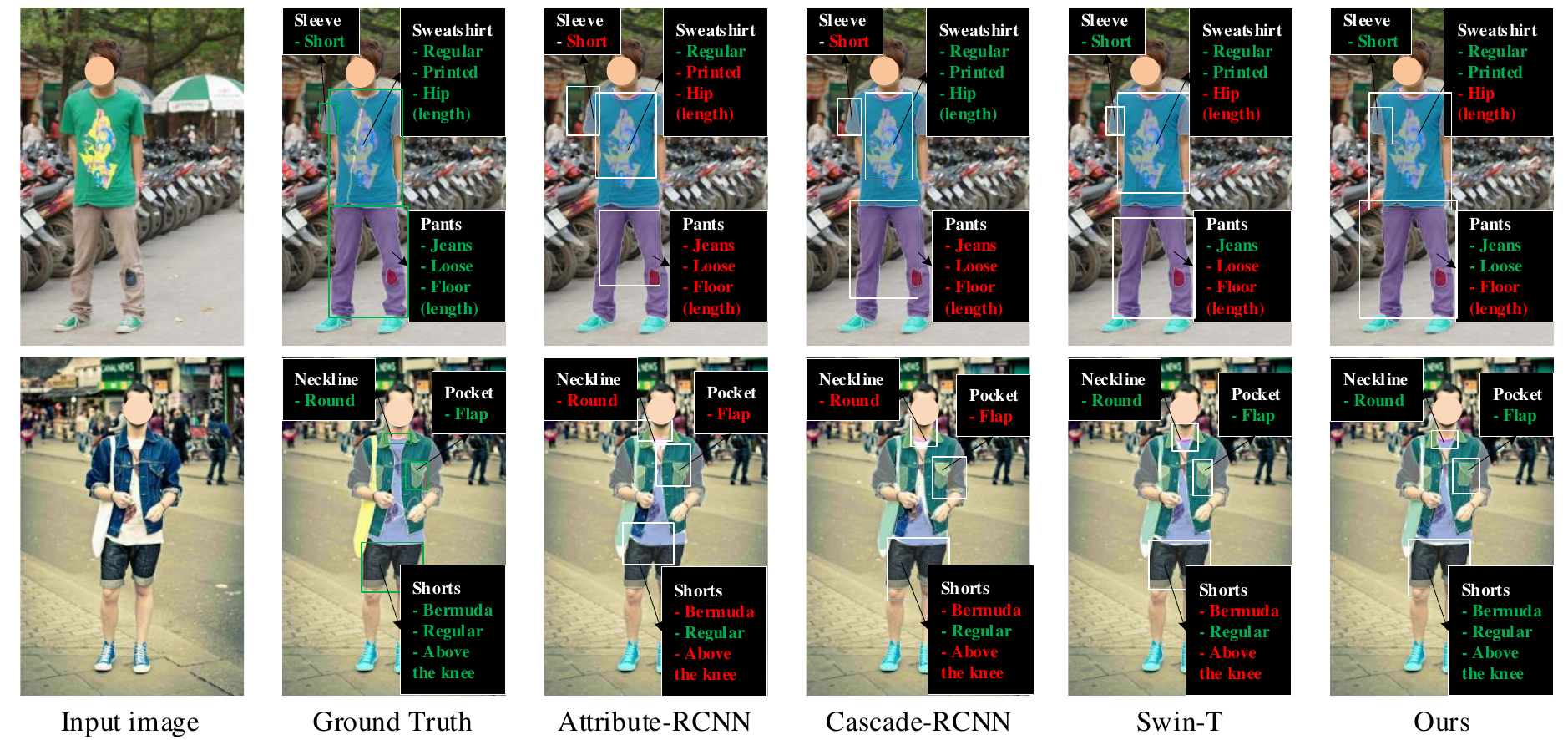} 
	\caption{Qualitative comparison. From the left to right: the input images, ground truth part-level attributes, predictions from Attribute-RCNN\cite{dataset:fashionpedia}, Cascade-RCNN\cite{obj_det:cascadedrcnn}, Swin-T \cite{swintrans} and ours, respectively. Color green denotes corrected attribute parsing by model and color red denotes failure cases. Each part is drawn in a distinguished color. Zoom in for a better view.}
	\label{fig:qua_com}
\end{figure*}
\noindent\textbf{Qualitative comparison.} In Fig.~\ref{fig:qua_com} we show some examples of qualitative part-level attribute parsing results randomly selected from the Fashionpedia dataset. It is clear that our method performs the best while the Attribute-RCNN performs the worst. In addition, given six parts examples, our method provides four parts with completely correct attribute predictions, while the most competitive method Swin-T provides three parts with completely right attributes. To summary, this qualitative results are consistent with the quantitative results shown in Tab.~\ref{tab:all_com_stoa_fashion}, which demonstrates the superiority of our proposed method and proves the important role of implicit and explicit knowledge in part-level attribute parsing. 

\noindent\textbf{The bottleneck of KE-RCNN.}
The part-level attribute parsing depends on the precise predictions of sub-tasks, i.e., the body detection, the body part detection and the prediction of attributes. In this section, we perform an experiment to gauge the relative difficulty of sub-tasks, that is, which part is the main bottleneck of part-level attribute parsing so far. We evaluate KE-RCNN on the Fashionpedia dataset as well as Kinetics-TPS dataset, hoping to inspire future research. Specifically, we replace predictions with corresponding ground-truth labels. The results in Tab.~\ref{tab:upper_bound_analysis} suggest that there is still a large room for improvement in part detection and attribute parsing.

\begin{figure*}[ht]
	\centering
	\includegraphics[width=0.75\linewidth]{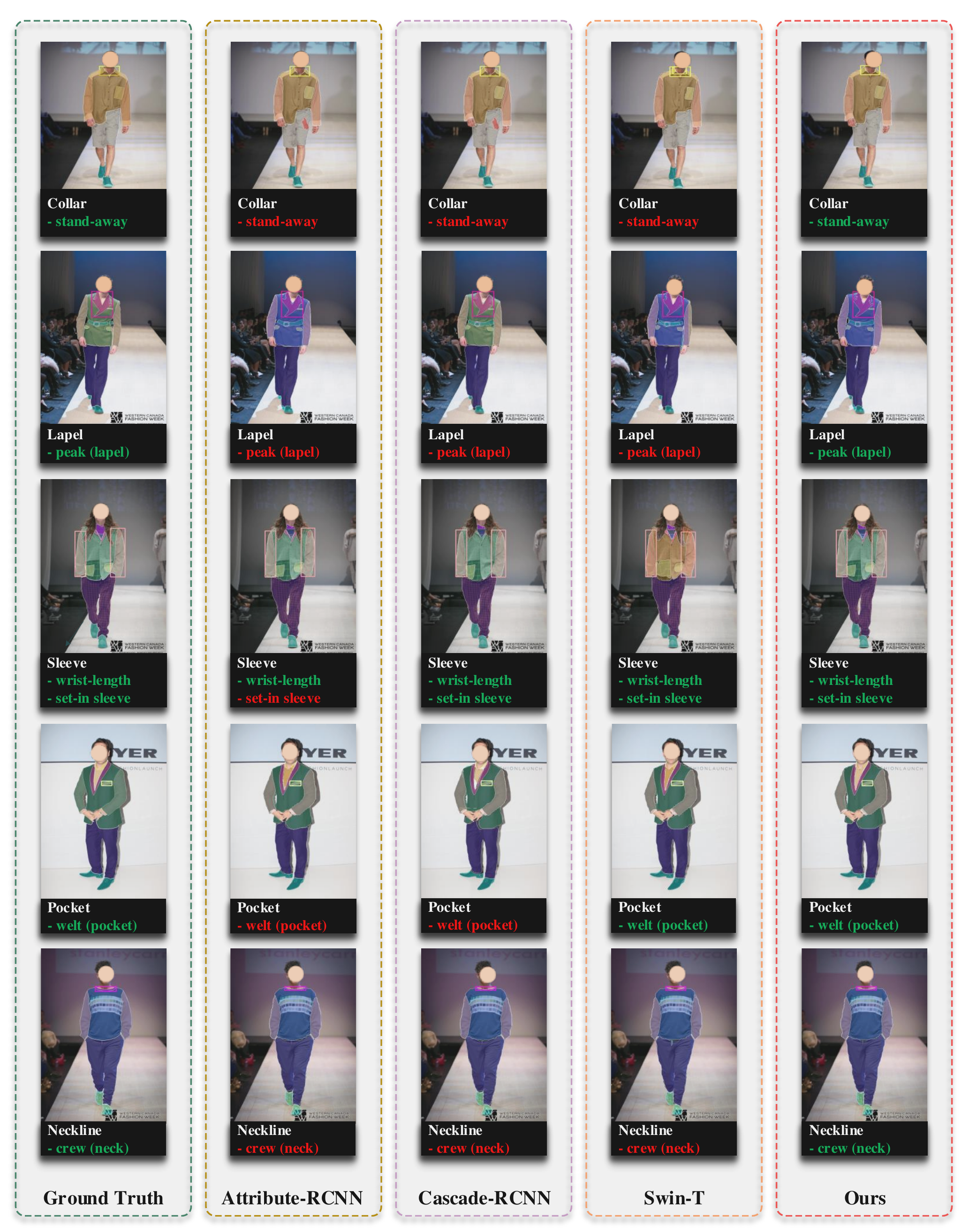}
	\centering
	\caption{Qualitative comparison for parsing ``hard'' cases, involving \textit{Collar}, \textit{Lapel}, \textit{Sleeve}, \textit{Pocket} and \textit{Neckline}. Color green denotes corrected attribute parsed by model and color red denotes failure cases.}
	\label{fig.qua_fewshot_comp_res}
\end{figure*}

In addition, Fig.~\ref{fig.qua_fewshot_comp_res} shows visualization results obtained from predictions for ``hard'' examples. From Fig.~\ref{fig.qua_fewshot_comp_res}, one can conclude that ``hard'' examples share a common characteristic of ``small''. Intuitively, parsing those hard examples requires rich visual contexts. The proposed KE-RCNN jointly encodes such contexts by implicit knowledge modeling as well as explicit knowledge modeling, thus improving overall performance for those ``hard'' examples. For more details, we refer the reader to supplementary materials. 

\begin{table*}[ht]
	\centering
	\caption{Upper bound analysis of part-level human attribute parsing by using ground-truth labels. The results suggest that there is still a large room for improvement in part detection and attribute parsing.}
	\renewcommand\arraystretch{1.2}
	\resizebox{0.9\linewidth}{!}{
		\begin{tabular}{|c|ccccc|}
			\hline
			\multicolumn{6}{|c|}{Part-leve Fashion Attribute Parsing on Fashionpedia Dataset}                                                                    \\ \hline
			Methods                & \multicolumn{1}{c|}{GT-$box_{part}$}   & $AP^{box}_{part}$ & $AP^{all}_{IoU+F_{1}}$  & $AP^{outerwear}_{IoU+F_{1}}$ &  $AP^{parts}_{IoU+F_{1}}$ \\ \hline
			\multirow{2}{*}{Attribute-RCNN} &    \multicolumn{1}{c|}{}             &    41.9      & 39.1\%             & 44.2\%           & 26.0\%           \\ \cline{2-6} 
			&         \multicolumn{1}{c|}{\checkmark}        & 99.9\% (\textcolor{red}{+58.0\%})   &   61.4\% (\textcolor{red}{+22.3\%})   & 66.7\% (\textcolor{red}{+22.5\%})       & 42.1\% (\textcolor{red}{+16.1\%})      \\ \hline
			\hline
			\multicolumn{6}{|c|}{Part-level Action Attribute Parsing on Kinetics-TPS Dataset}                                                                 \\ \hline 
			Methods                & GT-$box_{person}$ & \multicolumn{1}{c|}{GT-$box_{part}$}  & $AP^{box}_{person}$ & $AP^{box}_{part}$ & $Acc_{p}$       \\ \hline
			\multirow{3}{*}{Attribute-RCNN} &                 &    \multicolumn{1}{c|}{}            & 72.8\%             & 84.8\%           & 53.5\%           \\ \cline{2-6} 
			&      \checkmark           &         \multicolumn{1}{c|}{}       & 99.9\% (\textcolor{red}{+27.1\%})              &    86.3\%(\textcolor{red}{+1.8\%})     & 54.6\% (\textcolor{red}{+1.1\%})  \\ \cline{2-6} 
			&          \checkmark       &    \multicolumn{1}{c|}{\checkmark}          & 99.9\% (\textcolor{red}{+27.1\%})            & 99.9\% (\textcolor{red}{+15.1\%})          & 62.0\%  (\textcolor{red}{+8.5\%})        \\ \hline
		\end{tabular}
	}
	\label{tab:upper_bound_analysis}
\end{table*}

\section{Conclusion and Discussion}
In this paper, we propose an effective method named Knowledge Embedded RCNN (KE-RCNN), aiming at identifying part-level attributes by utilizing implicit and explicit knowledge. By building Implicit Knowledge based Encoder (IK-En), we enhance part representations by incorporating visual contexts as well as geometry contexts. Then Explicit Knowledge based Decoder (EK-De) is proposed to identify attributes of a part by human prior knowledge. Extensive experiments on two benchmarks prove the effectiveness and generalizability of our approach. 

\bibliographystyle{IEEEtran}
\bibliography{bib}

\section*{A1. Detailed Architecture}
\begin{figure}[t]
	\centering
	\setlength{\abovecaptionskip}{5.pt}%
	\includegraphics[width=1.\linewidth]{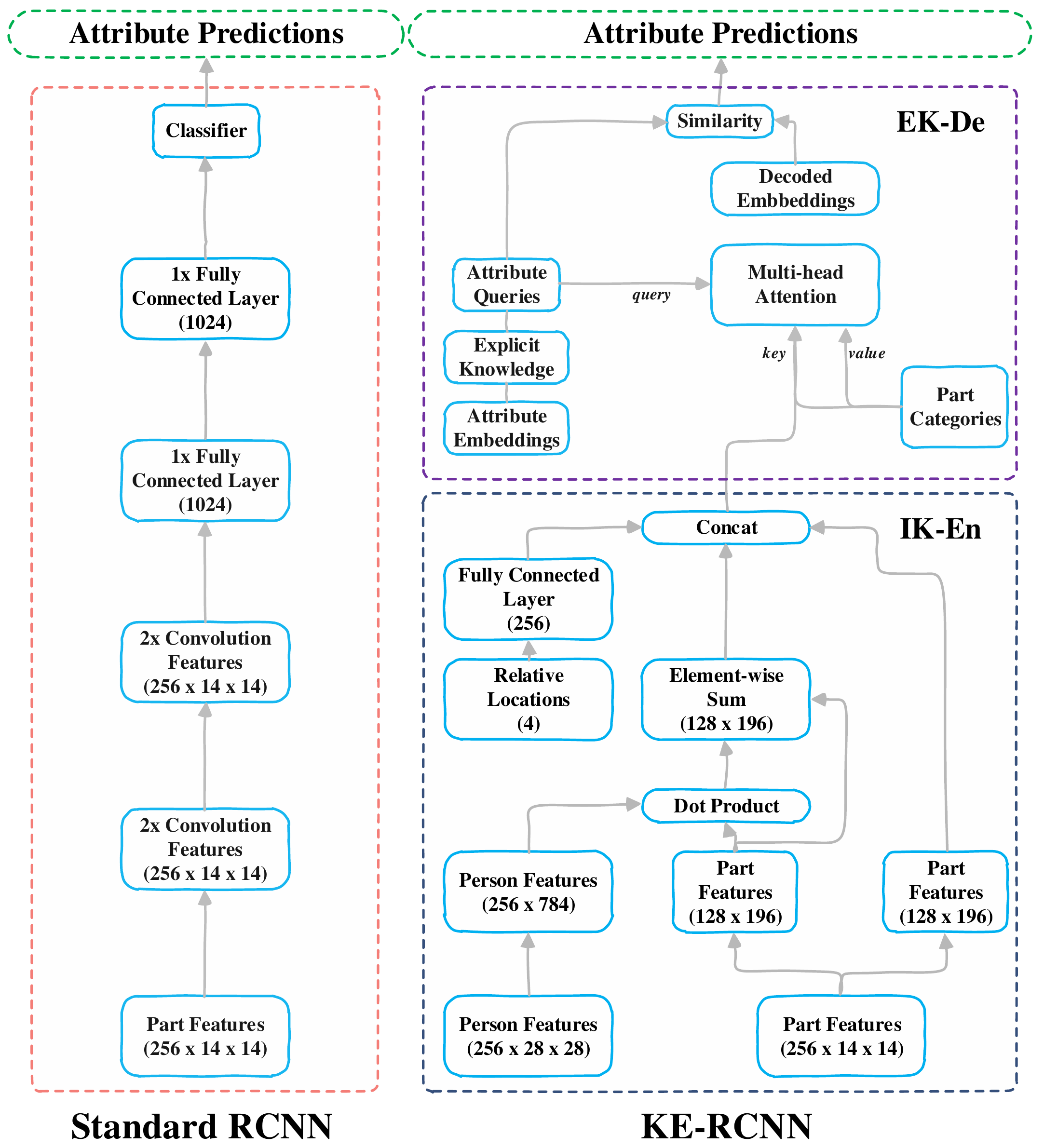}
	\centering
	\caption{Topological structure of attribute parsing branch for both standard RCNN (left) and KE-RCNN (right).}
	\label{FIG.struct}
	\vspace{-0.5cm}
\end{figure}

In terms of part-level attribute parsing branch, Fig.~\ref{FIG.struct} presents major differences between standard RCNN and proposed KE-RCNN. For each model, Tab.~\ref{Tab.modelsize} further lists the model size, FLOPs and $AP^{all}_{IoU+F_{1}}$. From the results, we can see that the parameter size of the standard RCNN is around 61.1M, 256.0M for ASPP RCNN and 48.9M for the KE-RCNN with ResNet-50 backbone. Thus, the size of KE-RCNN has 12.2M and 207.1M smaller respectively than that of standard RCNN and ASPP RCNN, but the performance of KE-RCNN has +4.0\% $AP^{all}_{IoU+F_{1}}$ and +3.3\% $AP^{all}_{IoU+F_{1}}$ larger than that of standard RCNN and ASPP RCNN. Furthermore, in terms of speed comparison, all KE-RCNN based models process around 10 frames per second (fps), which are comparable to baseline models. This implies that knowledge modeling is more important than increasing model size for improving ability of attribute parsing.
\begin{table}[ht]
	\centering
	\caption{\# Params and FLOPs of baseline models and KE-RCNN. The FLOPs is computed with one proposal. M=$10^6$,G=$2^{30}$}
	\renewcommand\arraystretch{1.2}
	\resizebox{0.99\linewidth}{!}{
		\begin{tabular}{|c|ccccc|}
			\hline
			Settings       & Bacbone   & \# Params                        & FLOPs        &  Speed (img/s)             & $AP^{all}_{IoU+F_{1}}$                                              \\ \hline
			Standard RCNN  & ResNet-50 &  61.1M                     &     391.3G        &   15.3      &   35.1                                                    \\ 
			ASPP RCNN  & ResNet-50 &  256.0M                     &     754.0G        &   14.5      &   35.8                                                    \\ \hline	
			KE-RCNN       &ResNet-50  &    48.9M                  & 445.3G        &    12.8          &   39.1                                                      \\ 
			KE-RCNN       &HRNet-18  &    32.2M                  & 373.4G        &    11.1          &   36.4                                                      \\ 
			KE-RCNN       &Swin-T  &    52.2M                  & 449.8G        &    9.5          &   42.1                                                      \\ 
			KE-RCNN       &Swin-S  &    75.5M                  & 542.5G        &    7.9          &   44.3                                                      \\ \hline
		\end{tabular}
	}
	\label{Tab.modelsize}
\end{table}

\begin{figure}[ht]
	\centering
	\includegraphics[width=0.99\linewidth]{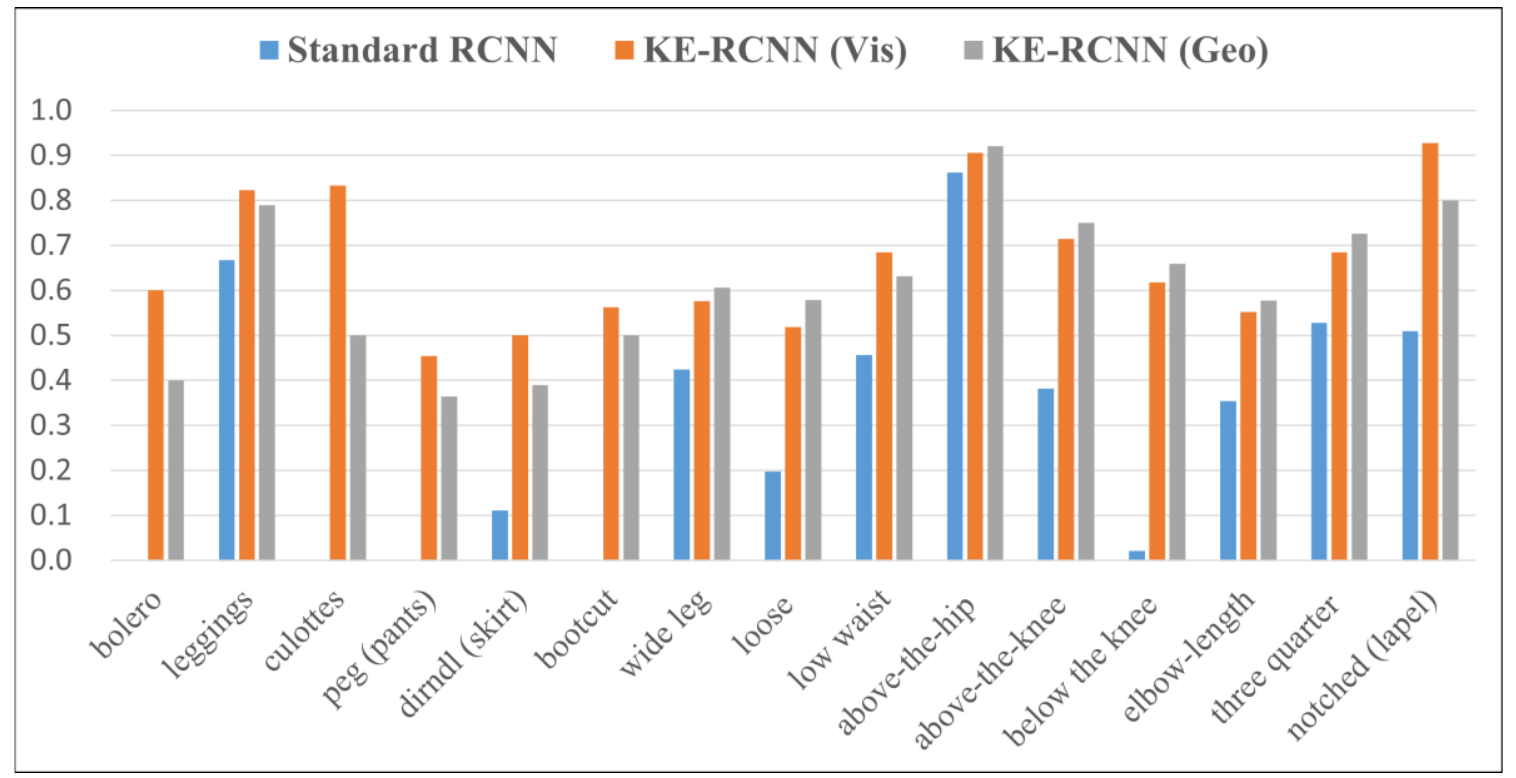}
	\caption{Category based average accuracy, which assesses Standard RCNN and two KE-RCNN methods with partial implicit knowledges (visual context or geometry context). All models are evaluated on Fashionpedia dataset. Only partial representative categories are listed due to the space limitation.}
	\label{fig.vis_geo_comp}
\end{figure}

\section*{A2. A Deep Analysis of Knowledge Modeling in KE-RCNN}

\subsection*{A2.1. Implicit Knowledge Modeling in IK-En.}
\begin{figure*}[t]
	\centering
	\includegraphics[width=0.9\linewidth,height=1.1\linewidth]{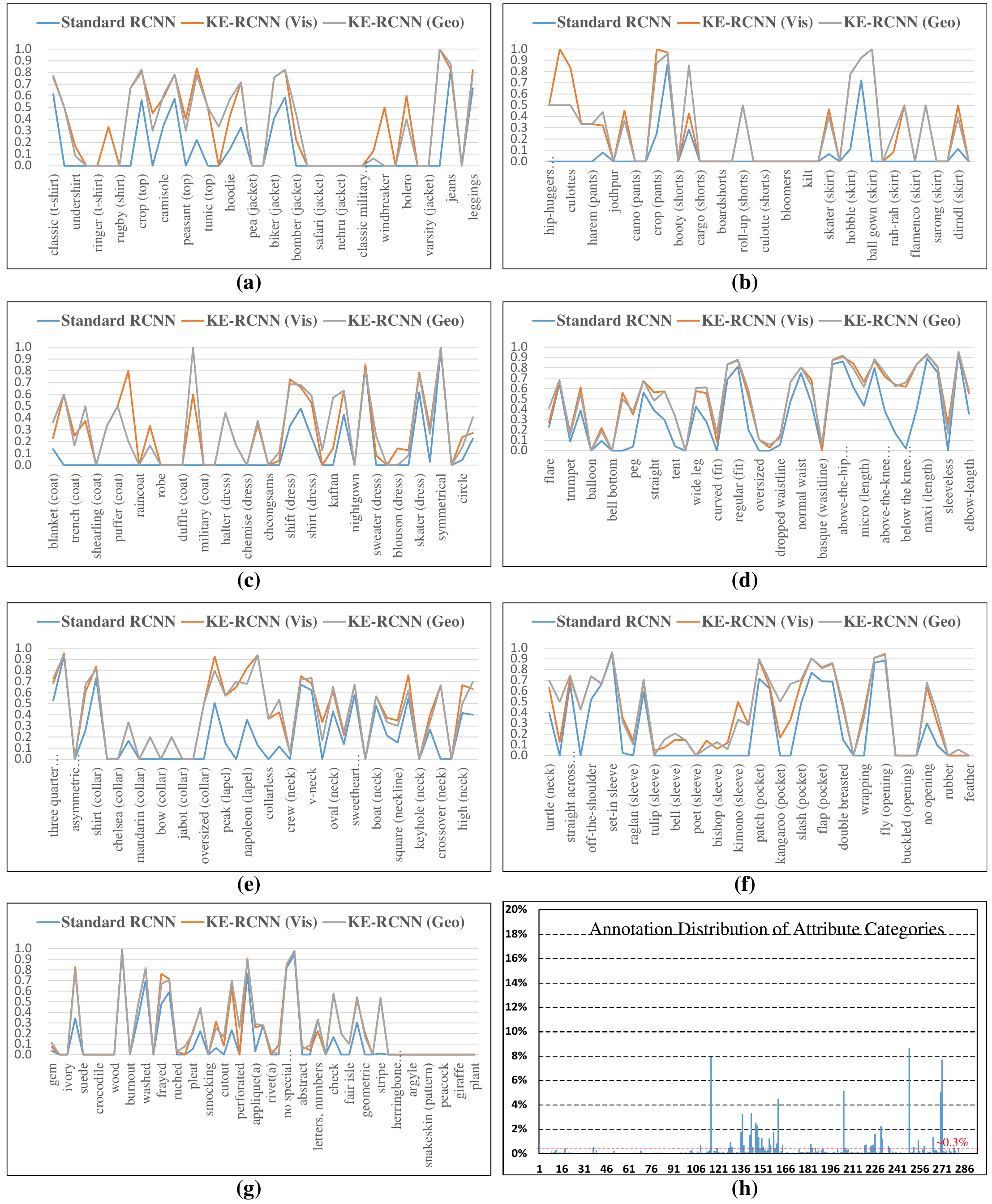}
	\centering
	\caption{Investigating what attributes benefit from implicit knowledge and which implicit knowledge they need to add, i.e., visual contexts or geometry contexts. Sub-figures (a)-(g) show accuracy curves of all attribute categories, where each sub-figure covers around 40 classes and all models are evaluated on Fashionpedia dataset. The sub-figure (h) indicates annotation distribution of attribute categories, which presents a characteristic of imbalance. In most of cases, KE-RCNN outperforms standard RCNN, where most attribute categories benefit from both visual context and geometry context.}
	\label{fig.full_comp_vis_geo}
\end{figure*}

The core idea of implicit knowledge modeling in IK-En lies in two global context encodings, where the IK-En jointly incorporates relevant visual contexts and geometry contexts into a part representation. Hence, one question is posed: how much each context contributes to identifying attributes. To answer this, Fig.~\ref{fig.vis_geo_comp} lists class-wise accuracy scores and compares three different models: 1) standard RCNN without implicit knowledge modeling; 2) the IK-En in KE-RCNN is only with visual context encoding; and 3) the IK-En in KE-RCNN is only with geometry context encoding. In particular, we use ground-truth bounding boxes as inputs, eliminating adverse effects caused by false part detection. From the results, we find KE-RCNN with visual context encoding improves standard RCNN on fix attributes, which are related to cloth patterns or cloth styles, such as ``dirndl'' and ``culottes''. As for those location-sensitive attributes, such as ``above-the-hip'' and ``three quarter length'', the KE-RCNN with geometry context encoding improves standard RCNN most. Therefore, one can conclude that both visual contexts and geometry contexts are indispensable for accurate attribute parsing. For more evaluation results, we refer readers to Fig.~\ref{fig.full_comp_vis_geo}.
\begin{figure*}[ht]
	\centering
	\includegraphics[width=0.99\linewidth]{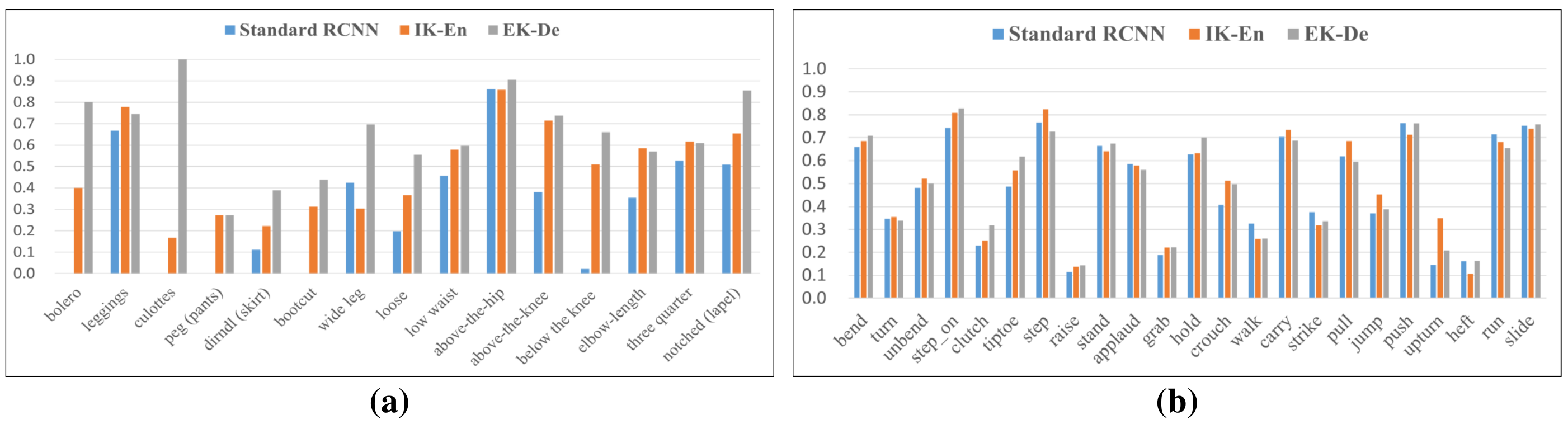}
	\centering
	\caption{Category based average accuracy, which assesses Standard RCNN, IK-En and EK-De. All models are evaluated on Fashionpedia dataset (a) Kinetics-TPS dataset (b). Only partial representative categories are listed due to the space limitation}
	\label{fig.ik-ek-comp}
\end{figure*}
\subsection*{A2.2. Explicit Knowledge Modeling in EK-De.}
Explicit knowledge can be summarized from different data sources, i.e., either from annotations of a dataset or from task-irrelevant information source. Hence, annotations of Fashionpedia and large scale language corpus in Wikipedia are used to construct explicit knowledges about part-attribute relations. Tab.~\ref{tab:abla_study_comp_knowledge_fashion} compares Fashionpedia and Wikipedia on part-level fashion parsing, where they are applied to KE-RCNN with different backbones. While the Fashionpedia is used as a default data source for explicit knowledge generation, we generally observe comparable accuracy by replacing it with the Wikipedia. It indicates that the difference between the two kinds of knowledge is slightly minor. We thus use annotations of Fashionpedia to construct explicit knowledge about part-attribute relations.

\begin{table}[t]	
	\centering
	\caption{Investigating the effect of prior knowledge.}
	\renewcommand\arraystretch{1.8}
	\resizebox{1.\linewidth}{!}{
		\begin{tabular}{|c|c|c|c|c|} 
			\hline
			Settings & Knowledge Source  &  $AP^{all}_{IoU+F_{1}}$   \\
			\hline
			\multirow{2}{*}{KE-RCNN (ResNet-50)} & Fashionpedia &  39.1   \\
			& Wikipedia & 39.6      \\
			\hline
			\multirow{2}{*}{KE-RCNN (ResNet-101)} & Fashionpedia & 39.9   \\
			& Wikipedia & 40.7      \\
			\hline
			\multirow{2}{*}{KE-RCNN (Cascade-R50)} & Fashionpedia &  41.2   \\
			& Wikipedia & 41.6     \\
			\hline
			\multirow{2}{*}{KE-RCNN (Cascade-R101)} & Fashionpedia & 42.7    \\
			& Wikipedia & 42.3      \\
			\hline
			\multirow{2}{*}{KE-RCNN (HRNet-18)} & Fashionpedia & 36.4   \\
			& Wikipedia & 37.7       \\
			\hline
			\multirow{2}{*}{KE-RCNN (HRNet-32)} & Fashionpedia & 39.0   \\
			& Wikipedia & 39.2     \\
			\hline
			\multirow{2}{*}{KE-RCNN (Swin-T)} & Fashionpedia & 42.1     \\
			& Wikipedia & 41.7      \\
			\hline
			\multirow{2}{*}{KE-RCNN (Swin-S)} & Fashionpedia & 44.3     \\
			& Wikipedia & 45.0   \\
			\hline
	\end{tabular}}
	\vspace{-0.5cm}
	\label{tab:abla_study_comp_knowledge_fashion}
\end{table}

\subsection*{A2.3. Implicit Knowledge or Explicit Knowledge ?}
In this section, we would like to investigate what attributes can benefit from knowledge modeling and which knowledge they prefer. To answer this,  we compare three different models: 1) a standard RCNN without knowledge modeling; 2) KE-RCNN with IK-En only; and 3) KE-RCNN with EK-De only. For each model, we conduct class-wise evaluation and calculate the attribute classification accuracy. The corresponding accuracy scores are summarized in Fig.~\ref{fig.ik-ek-comp}, where only partial representative categories are presented due to the space limitation. From the results, we generally observe improved accuracy for most attributes after applying proposed IK-En or EK-De, indicating general effectiveness of proposed method. 

In terms of static attributes (i.e., fashion), as shown in Fig.~\ref{fig.ik-ek-comp} (a), most attributes prefer EK-De as it generally outperforms other two approaches by a large margin. We conjecture the possible reason behind this is that EK-De significantly narrows down the recognition space via statistical priors since fashion attributes are strongly correlated to corresponding clothes. In this way, the attribute parsing model only focus on few attributes, thus benefiting final parsing performance.  
As for dynamic attributes (i.e., action states), as shown in Fig.~\ref{fig.ik-ek-comp} (b), most attributes with diverse poses or large motion changes, such as ``jump'' and ``upturn'', generally benefit more from IK-En than EK-De. However, those attributes with a fixed pose or slight motion variations, such as ``clutch'' and ``hold'', prefer EK-De. Thus, one can conclude that EK-De is particularly helpful for those attributes with a strong correlation to the body part or those with slight changes, while IK-En is generally beneficial to those complex attributes with large changes or location sensitive ones. For more details, we refer readers to Fig.~\ref{fig.full_comp_ik_ek_fashion} and Fig.~\ref{fig.full_comp_ik_ek_kinetics}.

\begin{table}[ht]
	\centering
	\caption{Ablation study on Fashionpedia. Investigating the effect of training practices.}
	\renewcommand\arraystretch{1.2}
	\resizebox{0.75\linewidth}{!}{
		\begin{tabular}{|c|c|c|c|c|}
			\hline
			Initialization & Epoch & $Acc_{p}$ & $Acc_{s}$   & $AP^{box}_{part}$ \\ \hline
			ImageNet       & 24    &    50.6       &   65.7          &    84.9               \\ \hline
			COCO           & 12    &    51.2       &   66.7          &    84.9               \\ \hline
			COCO           & 24    &    52.7       &   68.5          &    84.7               \\ \hline
		\end{tabular}	
	}
	\label{tab:abla_study_train_prac_kinetics}
\end{table}

\begin{table}[ht]
	\centering
	\caption{Kinetics-TPS Challenge results on \textit{test} set. Comparison results are directly summarized from DeeperAction competition page.}
	\renewcommand\arraystretch{1.3}
	\resizebox{0.65\linewidth}{!}{
		\begin{tabular}{|c|c|c|}
			\hline
			\multicolumn{3}{|c|}{2021 Leaderboard} \\ \hline 
			Ranks &	Participants          & ${Acc}_{p}$                 \\ \hline
			(1) &	yuzheming       & 0.630532           \\ 
			(2)	&Sheldong        & 0.613722           \\ 
			(3)	& JosonChan   & 0.605059  \\ 
			(4) 	&fangwudi        & 0.590167          \\ 
			(5)	&uestc.wxh       & 0.536067           \\
			(6)	&hubincsu        & 0.490984           \\ 
			(6) 	&scc1997         & 0.490984          \\ 
			(7)	&KGH             & 0.486483           \\ 
			(8)	&zhao\_THU        & 0.434311           \\ 
			(9)	&TerminusBazinga & 0.396735           \\ 
			(10)	&cjx\_AILab       & 0.370753          \\ 
			(11)	&xubocheng       & 0.358834          \\ 
			(12)	&haifwu          & 0.247614          \\ 
			(13)	&Aicity          & 0.189669          \\ 
			(14)	&fog             & 0.188455          \\ \hline
			- & KE-RCNN (Ours) &  0.653541  \\ \hline
		\end{tabular}
	}
	\label{tab.challenge_test}
\end{table}
\section*{A3. Detailed Training Practices on Kinetics-TPS}
This section provides training details of part-level action parsing models. In particular, we consider two training factors, including initialization method and learning schedule. In general, a good initialization will lead to better results. In line with this finding, we observe that action parsing models pre-trained on COCO dataset outperforms the counterpart that is pre-trained on ImageNet dataset, as shown in Tab.\ref{tab:abla_study_train_prac_kinetics}. Besides, Tab.~\ref{tab:abla_study_train_prac_kinetics} also compares $1\times$ training schedule and $2\times$ training schedule, where it indicates longer training schedule will be likely to obtain a better parsing model. 

\noindent\textbf{DeeperAction 2022 Challenge:} After exploring good practices of training action parsing models, we participated in the DeeperAction 2022 Kinetics-TPS track. Our proposed method outperforms all entries from the leaderboard of DeeperAction 2021 Kinetics-TPS, as shown in Tab.~\ref{tab.challenge_test}. Our entry only uses a single model of KE-RCNN with Swin-B backbone, and attains a score of 65\% on the test server, which surpasses the 1st place in 2021 leaderboard by 2.3 points. 

\section*{A4. The Limitation of KE-RCNN}
\begin{figure*}[tt]
	\centering
	\includegraphics[width=0.9\linewidth,height=1.2\linewidth]{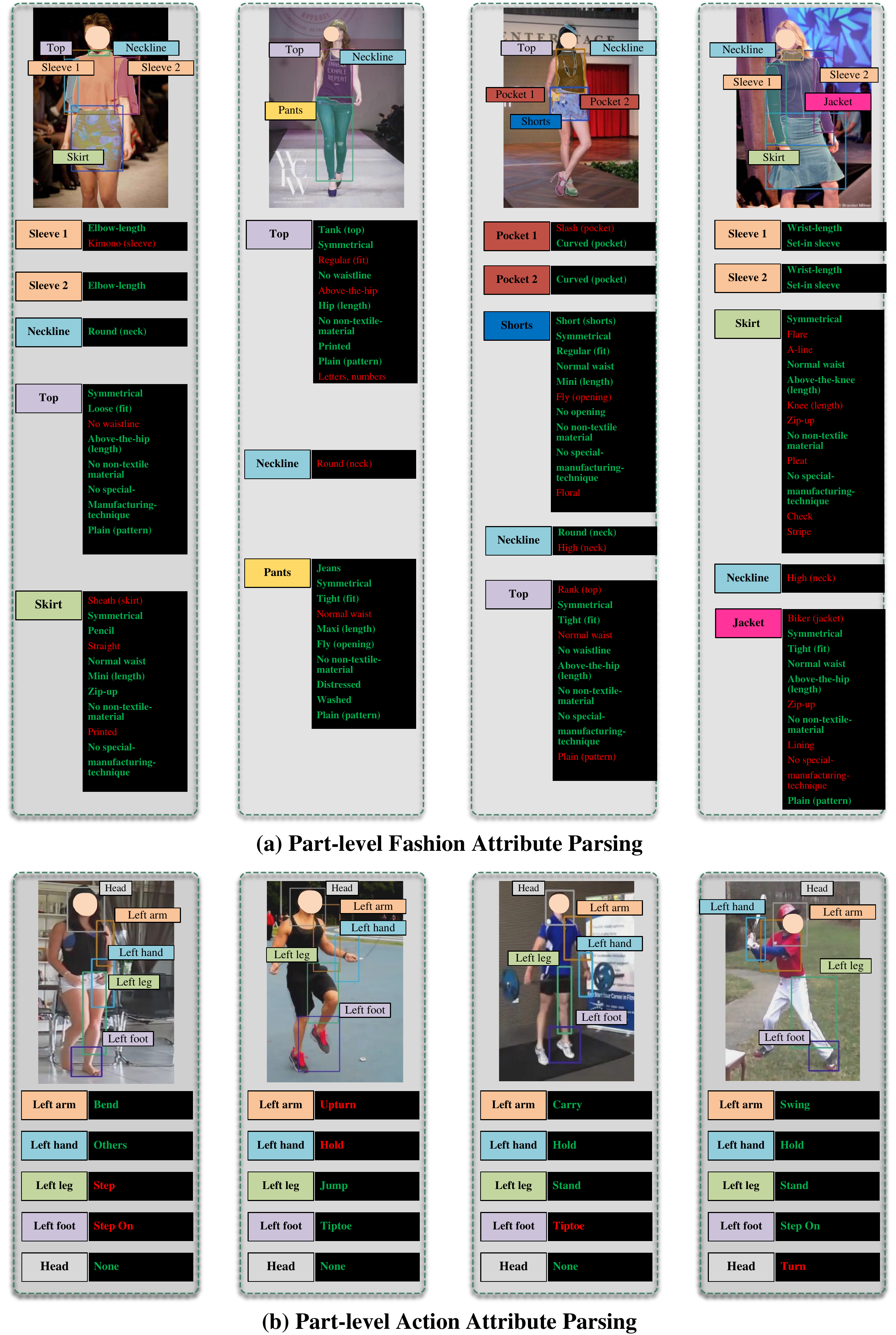}
	\centering
	\caption{Visualization results obtained from KE-RCNN on Fashionpedia dataset (a) and Kinetics-TPS dataset (b). Color green denotes corrected attribute parsing by model and color red denotes failure cases. Each part is drawn in a distinguished color. Zoom in for a better view.}
	\label{fig.qua_res}
\end{figure*}

\begin{figure*}[t]
	\centering
	\includegraphics[width=0.9\linewidth,height=1.1\linewidth]{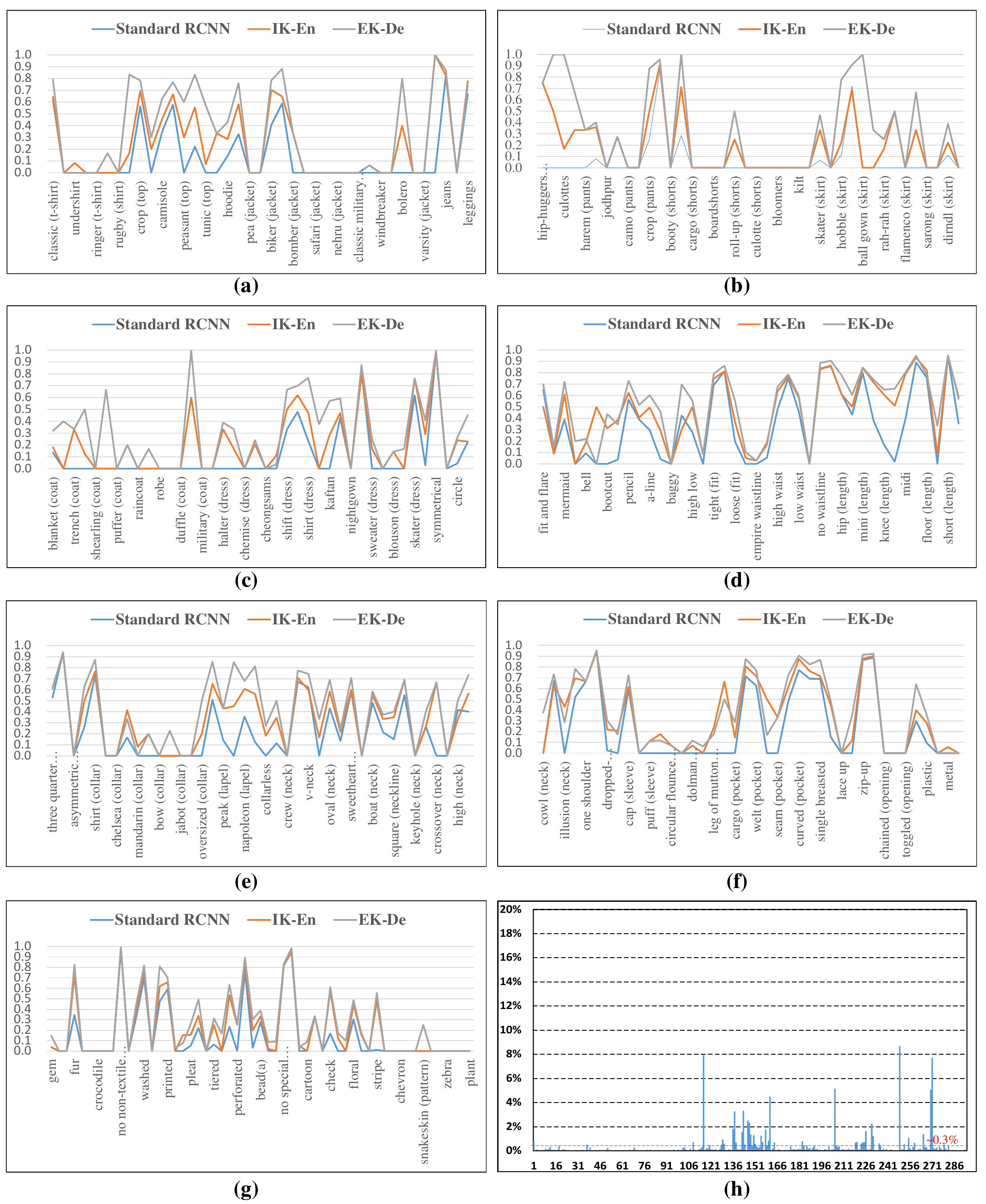}
	\centering
	\caption{Investigating what static attributes benefit from knowledge modeling and which knowledge they need to add, i.e., implicit knowledge or explicit knowledge. Sub-figures (a)-(g) show accuracy curves of all attribute categories, where each sub-figure covers around 40 classes and all models are evaluated on Fashionpedia dataset. In most of cases, KE-RCNN outperforms standard RCNN, where most attribute categories benefit from both implicit and explicit knowledge.}
	\label{fig.full_comp_ik_ek_fashion}
\end{figure*}

\begin{figure*}[t]
	\centering
	\includegraphics[width=0.95\linewidth]{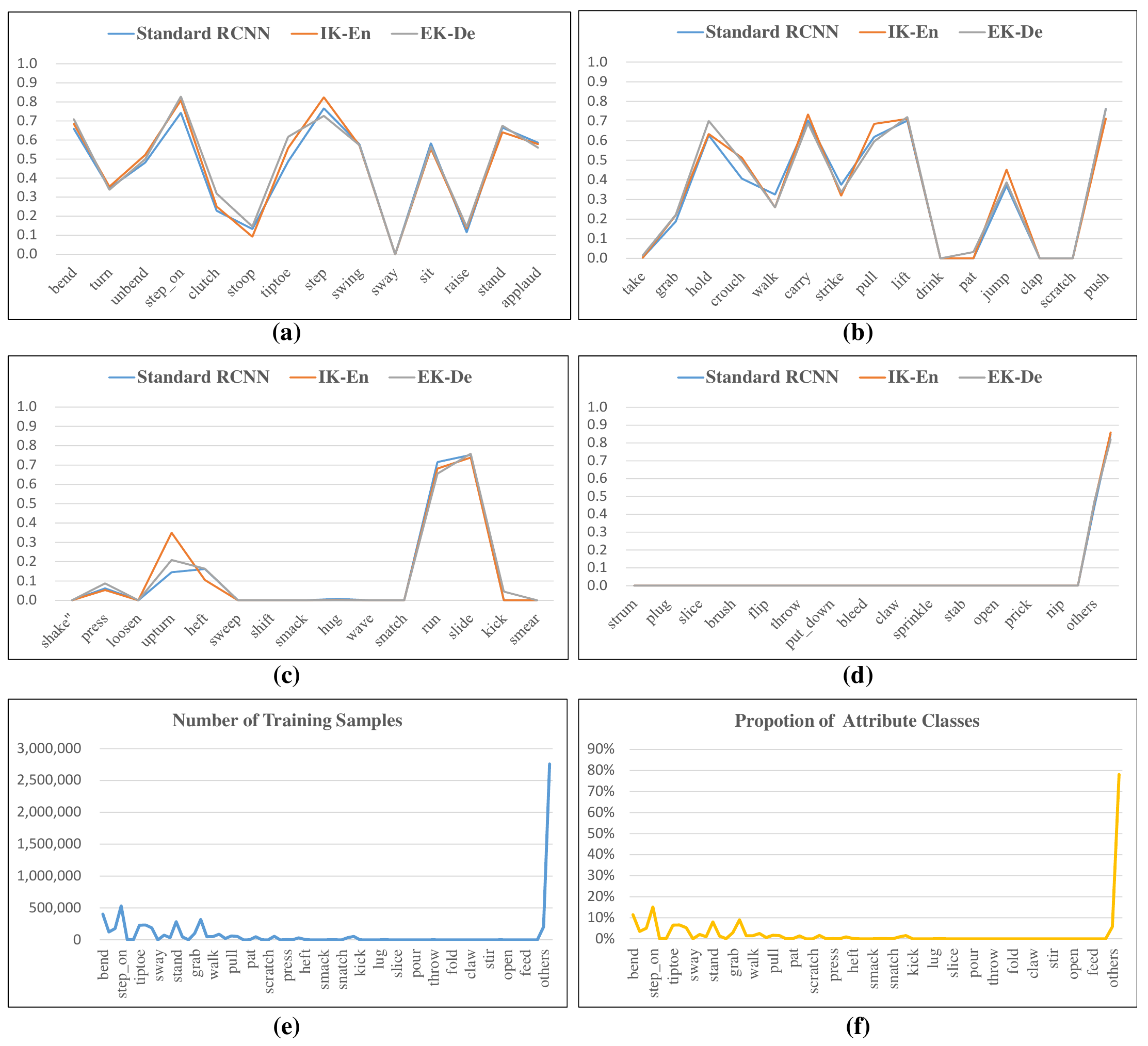}
	\centering
	\caption{Investigating what dynamic attributes benefit from knowledge modeling and which knowledge they need to add, i.e., implicit knowledge or explicit knowledge. Sub-figures (a)-(d) show accuracy curves of all attribute categories, where each sub-figure covers around 15 classes and all models are evaluated on Kinetics-TPS dataset. The sub-figure (e)-(f) indicate annotation distribution of attribute categories, where many classes involve very few samples. Most attribute categories benefit from both implicit and explicit knowledge.}
	\label{fig.full_comp_ik_ek_kinetics}
\end{figure*}

In this section, we discuss the limitation of the KE-RCNN. One major limitation is that the KE-RCNN highly depends on the part detection results. In this way, the performance of attribute parsing will drop significantly if the detection performance decreases dramatically. Fig.~\ref{fig.qua_res} shows some failure cases performed by KE-RCNN. Most failure cases suffer from misclassifying complex attributes conditioning on false predicted bounding boxes. The second limitation is that the number of labeled data is not sufficient. Both Fashionpedia and Kinetics-TPS present a characteristic of imbalance and sparse, as shown in Fig.\ref{fig.full_comp_ik_ek_fashion} and Fig.\ref{fig.full_comp_ik_ek_kinetics}. In particular, numerous action attribute classes in Kinetics-TPS dataset involve few training samples and most samples (around 78\%) are annotated as unknown state (i.e., labeled as ``none''). Therefore, the KE-RCNN directly trained on Kinetics-TPS performs worst on those minor classes, as shown in Fig.~\ref{fig.qua_res} and Fig.\ref{fig.full_comp_ik_ek_kinetics}.

To remedy for it, one possible solution is that parsing attributes conditioning on bounding box as well as mask for a part. The box decides rough location of a part and the mask decides which area the model should attend. Another possible solution is to design a standalone pipeline, where the model directly parsing instance-level attribute parsing results without extra part detection. In particular, the second one is the promising future research direction as single-stage recognition pipelines have been explored on many related research works, e.g., general object detection. As for the second limitation, one-shot or few-shot learning paradigm should be a good way for enhancing generalizability, which can be studied in the future works. For more details, we refer the reader to our project page\footnote{\url{https://github.com/sota-joson/KE-RCNN}}.

\clearpage

\clearpage
\end{document}